\documentclass{article}
\usepackage{pifont}
\usepackage{paralist}
\usepackage{subcaption}
\usepackage[final]{neurips_2025}
\usepackage{url}
\usepackage{amsmath}
\usepackage[utf8]{inputenc} 
\usepackage[T1]{fontenc}    
\usepackage{url}            
\usepackage{tabularx}
\usepackage{booktabs}
\usepackage{multirow}
\usepackage{amsfonts}       
\usepackage{nicefrac}       
\usepackage{microtype}      
\usepackage{natbib}
\usepackage[dvipsnames,table,xcdraw]{xcolor}
\definecolor{cvprblue}{rgb}{0.21,0.49,0.74}
\usepackage{graphicx}
\usepackage{caption}
\usepackage{arydshln}
\usepackage{tikz}
\usepackage{wrapfig}
\usepackage{float}
\usepackage{multirow}
\usepackage{pifont}
\usepackage{makecell}
\usepackage[pagebackref,breaklinks,colorlinks,citecolor=cvprblue]{hyperref}

\usepackage{mdframed}
\definecolor{mygray}{gray}{0.97}
\colorlet{shadecolor}{mygray}
\newmdenv[%
backgroundcolor=mygray, 
linewidth=0pt
]{newshaded}

\title{CSVQA: A Chinese Multimodal Benchmark for Evaluating STEM Reasoning Capabilities of VLMs}

\author{
	\textnormal{Ai Jian}\thanks{Equal contribution}, \quad \textnormal{Weijie Qiu}\footnotemark[1], \quad   \textnormal{Xiaokun Wang}, \quad   
	 \textnormal{Peiyu Wang},\quad 
	 \textnormal{Yunzhuo Hao},\\ \quad
	 \textnormal{Jiangbo Pei},\quad
	 \textnormal{Yichen Wei},\quad
	 \textnormal{Yi Peng},\quad
\textnormal{Xuchen Song}\thanks{Corresponding author}
    \\
\\
\textnormal{Skywork AI,  Kunlun Inc.}\\
{\textnormal{xuchen.song@kunlun-inc.com}}\\
}

\makeatletter
\newcommand{\@trackname}{} 
\makeatother
\begin{document}
		\maketitle
		\begin{abstract}
		Vision-Language Models (VLMs) have demonstrated remarkable progress in multimodal understanding, yet their capabilities for scientific reasoning remains inadequately assessed. Current multimodal benchmarks predominantly evaluate generic image comprehension or text-driven reasoning, lacking authentic scientific contexts that require domain-specific knowledge integration with visual evidence analysis. To fill this gap, we present CSVQA, a diagnostic multimodal benchmark specifically designed for evaluating scientific reasoning through domain-grounded visual question answering.
		Our benchmark features 1,378 carefully constructed question-answer pairs spanning diverse STEM disciplines, each demanding  domain knowledge,  integration of visual evidence, and higher-order reasoning. 
		Compared to prior multimodal benchmarks, CSVQA places greater emphasis on real-world scientific content and  complex reasoning.
		We additionally propose a rigorous evaluation protocol to systematically assess whether model predictions are substantiated by valid intermediate reasoning steps based on curated explanations. 
		Our comprehensive evaluation of 15 VLMs on this benchmark reveals notable performance disparities, as even the top-ranked  proprietary model attains only 49.6\% accuracy.
		This empirical evidence underscores the pressing need for advancing scientific reasoning capabilities in VLMs. Our CSVQA is released at \url{https://huggingface.co/datasets/Skywork/CSVQA}.
	\end{abstract}

	\section{Introduction}
	
	The emergence of large language models (LLMs) \cite{openai2023gpt4,llama2,deepseekr1} has significantly advanced natural language understanding and generation, paving the way for vision-language models (VLMs) that integrate visual and textual modalities for robust multimodal reasoning \cite{Qwen2VL,chen2023internvl,kimi1.5,geminiteam2024,openai2024gpt4o}. 
	As VLMs rapidly evolve, there is a growing need for multimodal benchmarks capable of rigorously evaluating their cross-modal comprehension and reasoning abilities.

While recent multimodal benchmarks have driven significant progress, they remain largely general-purpose, primarily focusing on everyday images and commonsense reasoning.
 State-of-the-art (SOTA) models such as InternVL2.5-78B~\cite{internvl2.5} have achieved impressive results, scoring 95.1\% on DocVQA~\cite{docvqa}, 84.1\% on InfoVQA~\cite{infovqa}, and 88.3\% on MMBench~\cite{MMB}. However, these benchmarks primarily assess perceptual understanding and fall short of rigorously evaluating the complex reasoning skills required in scientific domains.

	\begin{figure*}
		\centering
		\includegraphics[width=0.85\linewidth, height=0.35\textheight]{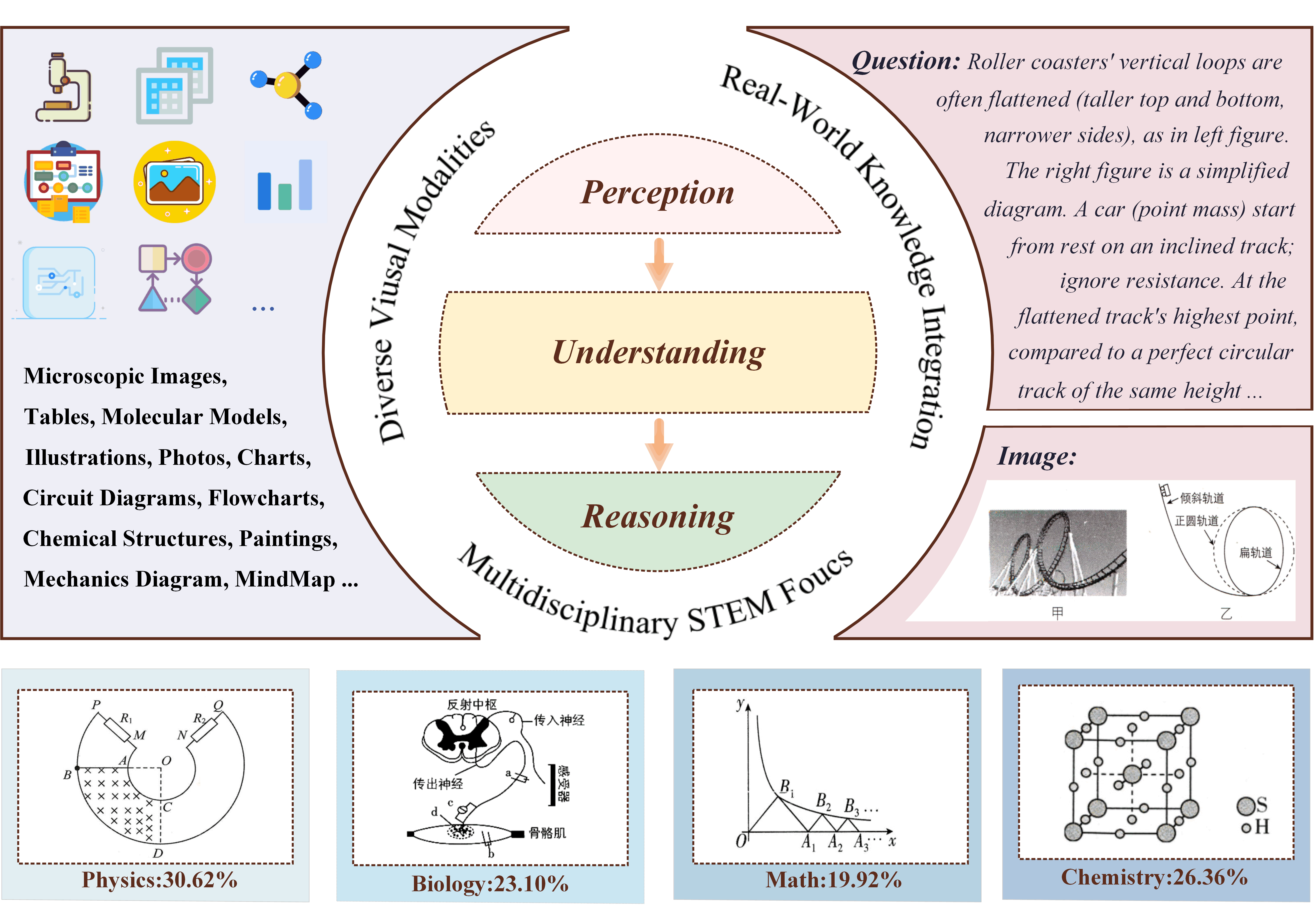}
		\caption{Challenges in the CSVQA Benchmark. (a) \textbf{Multidisciplinary STEM Focus}: Questions cover STEM subjects, ensuring comprehensive evaluation across disciplines; (b) \textbf{Diverse Visual Modalities}: The benchmark includes 14 types of images; (c) \textbf{Integration of Domain-Specific Knowledge and Real-World Scenarios}.}
		\label{challenges} 
		\vspace{-0.5em}
	\end{figure*}
	In this work, we address the need for a comprehensive evaluation of scientific reasoning in VLMs by introducing CSVQA, a human-aligned multimodal benchmark in Chinese contexts specifically crafted. 
	CSVQA is designed to evaluate the capability of VLMs to comprehend and reason about scientific content in images, moving beyond superficial pattern recognition to require integrated perception, understanding, and reasoning.
	Derived from authentic pedagogical materials across physics, chemistry, biology, and mathematics, the dataset comprises 1,378 rigorously validated problems with structured solution frameworks. 
	It features a multimodal design that incorporates 14 types of scientific visual modalities, supporting both multiple-choice diagnostics and open-ended reasoning tasks, which fills a critical evaluation gap and offers a more demanding testbed for assessing advanced reasoning capabilities. 
	Additionally, a challenging subset CSVQA-Hard is constructed based on difficulty scores and dependence on visual reasoning.
	
	We evaluate 15 VLMs including both open-source and closed-source models on CSVQA and observe three key findings. 
	These findings demonstrate that current VLMs still struggle with rigorous scientific reasoning:
 \begin{itemize}
 	\item \textbf{\textit{Substantial Performance Gap.} }
 	The best-performing model, o1~\cite{o1}, achieves only 49.6\% overall accuracy, highlighting persistent domain-specific challenges where even SOTA models achieve sub-60\% accuracy.
 	In the meanwhile, Qwen-2.5-78B-Instruct attains 38.5\% accuracy, the narrow gap between open-weight and proprietary systems revealing tangible progress in open model development,
 	yet underscores the significant challenges in scientific visual reasoning faced by current models.
 	\item \textbf{\textit{Modality Specific Weaknesses.}} Models exhibit pronounced deficiencies in physics-related tasks, with a notable drop in accuracy, likely attributable to the abstract and symbolic nature of visual representations. 
 	Similarly, performance on open-ended questions is substantially lower, indicating difficulties in generating coherent reasoning without predefined options. 
 	On the CSVQA-hard subset, accuracy further declines, exposing limitations in handling complex scientific concepts and visual reasoning abilities.
 	
 	\item \textbf{\textit{Reasoning Demand.}} Most models exhibit improved performance with Chain-of-Thought (CoT) prompting on our benchmark, indicating that the dataset comprises reasoning-intensive tasks requiring step-by-step logical inference.
 	To rigorously evaluate reasoning ability while minimizing the influence of random guessing, we introduce process-tracing experiments. 
 	The further performance drop observed in certain models highlights the gap between superficial pattern recognition and genuine multimodal reasoning.
 \end{itemize}

	\section{Related~Work}
	
	\textbf{Vision-Language Models.}
	Recent advances in large-scale multimodal pretraining have aimed at unifying vision and language representations, significantly improving the performance of VLMs. Early methods, such as LXMERT~\cite{tan2019lxmert} and UNITER \cite{chen2020uniter}, employed region-based features from object detectors and fused them with text encoders. 
	More recent models like CLIP \cite{radford2021learning} and ALIGN \cite{jia2021scaling} leverage large-scale image-text pairs sourced and use Vision Transformers to learn visual representations end-to-end. 
	Encoder-free models~ \cite{evev2,monointernvl} present an alternative approach where visual and textual information are integrated into a shared embedding space, eliminating the need for separate encoders for each modality.
	More recently, cutting-edge models such as Gemini 2.0~\cite{gemini} and o1~\cite{o1} have pushed the boundaries of multimodal reasoning by offering real-time perception, higher visual fidelity, and improved alignment across modalities. 
	
	\textbf{Commonsense VQA Benchmarks.}
	 Visual Question Answering (VQA) is a widely studied task that requires an AI system to answer questions based on image content.
	Early VQA datasets focus primarily on object recognition and general visual question answering~\cite{ai2d}. 
	Subsequent developments introduced broader capabilities spanning optical character recognition (OCR)~\cite{docvqa,infovqa,ocrvqa} and commonsense reasoning~\cite{MMB}.
	Knowledge-based VQA benchmarks have been proposed to push models beyond image content alone. 
	OK-VQA~\cite{marino2019okvqa} target questions that require external knowledge not evident in the image, such as world facts or commonsense. 
	While these require integrating information beyond the visual, the knowledge is typically of a general nature and not specialized scientific reasoning.

	\textbf{Multimodal Scientific Benchmarks.}
	Our work is closely related to benchmarks that combine images with scientific or educational question answering.
	One prominent example is ScienceQA~\cite{scienceqa}, which contains multiple-choice science questions spanning topics in natural science, social science, and language science, primarily at the elementary and high school level.
	However, ScienceQA’s many questions can be answered by recalling facts or reading text in the associated diagram, rather than truly interpreting novel visual data.
	To evaluate more advanced and diverse skills, several recent benchmarks extend multimodal QA to higher levels.
	MMMU~\cite{yue2024mmmu} is a benchmark with over 11k questions covering college-level knowledge across 30 subjects and EMMA~\cite{emma} is designed to evaluate visual-relied reasoning, however both of them offering little insight into the reasoning process of models.
	Progress in scientific domains has been mostly limited to mathematics, as seen in MathVista~\cite{mathvista} and MathVision~\cite{mathvision}.
	In contrast, CSVQA offers naturally occurring multimodal questions in Chinese, spanning multiple disciplines and accompanied by high rates of detailed step-by-step explanations, thereby providing a new litmus test for the reasoning abilities of VLMs.
		\begin{figure*}[htbp]
		\centering
		\includegraphics[width=1\linewidth]{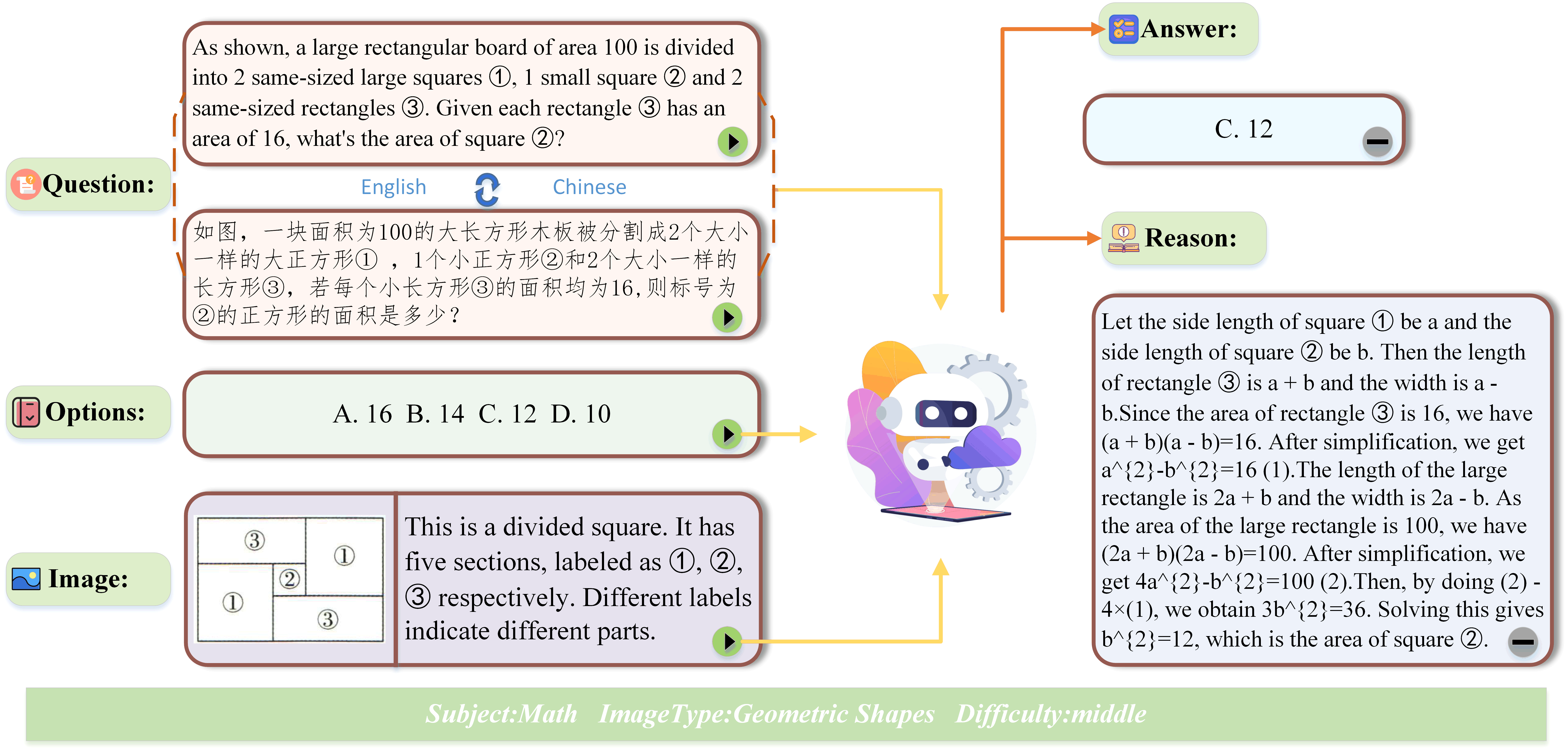}
		\caption{An example of a math problem from the CSVQA benchmark. Multi-step reasoning is required for solving the problem.} 
	\end{figure*} 
	\section{The~CSVQA~Benchmark}
	\subsection{Dataset Description}
	The Chinese STEM Visual Question Answering (CSVQA) benchmark establishes a rigorous multimodal evaluation framework specifically designed to quantify scientific reasoning capabilities in VLMs.
	Meanwhile, CSVQA introduces three key challenges that differentiate it from most existing benchmarks, as illustrated in Fig.~\ref{challenges}. 
	First, its coverage of multiple STEM disciplines which requires diverse domain knowledge and reasoning strategies. 
	Second, the inclusion of 14 distinct visual modalities introduces significant variation in visual structure and complexity, testing a model’s ability to generalize across image types. 
	Third, many questions are grounded in real-world scenarios and demand domain-specific knowledge, requiring models to go beyond pattern recognition and engage in context-aware reasoning.
	
\begin{figure}[htbp]
	\centering
	\begin{minipage}{0.48\textwidth}
		\centering
		\setlength{\tabcolsep}{3.5pt}
		\begin{tabular}{@{}ll@{}}
			\toprule
			\textbf{Statistics} & \textbf{Number} \\
			\midrule
			Total Questions & 1378\\
			Image Types & 14 \\
			Easy: Medium: Hard & 22.6\%:67.4\%:10.0\%\\
			\midrule
			Multiple-choice Question & 1278  \\
			Open Question & 100  \\
			\midrule
			With an Explanation & 81.1\%  \\
			\midrule
			Image in the Question & 1341  \\
			Image in Option & 37 \\
			\midrule
			Average question length & 69.7 \\
			Average option length & 12.1 \\
			Average explanation length & 123.5 \\
			\bottomrule
		\end{tabular}
		\captionof{table}{Key statistics of the CSVQA.\small\textcolor{gray}{To better compare with other datasets, the length analysis is conducted in English.}}
		\label{statistics}
	\end{minipage}%
	\hspace{0.3cm} 
	\begin{minipage}{0.48\textwidth}
		\centering
		\includegraphics[width=\textwidth,height=0.27\textheight]{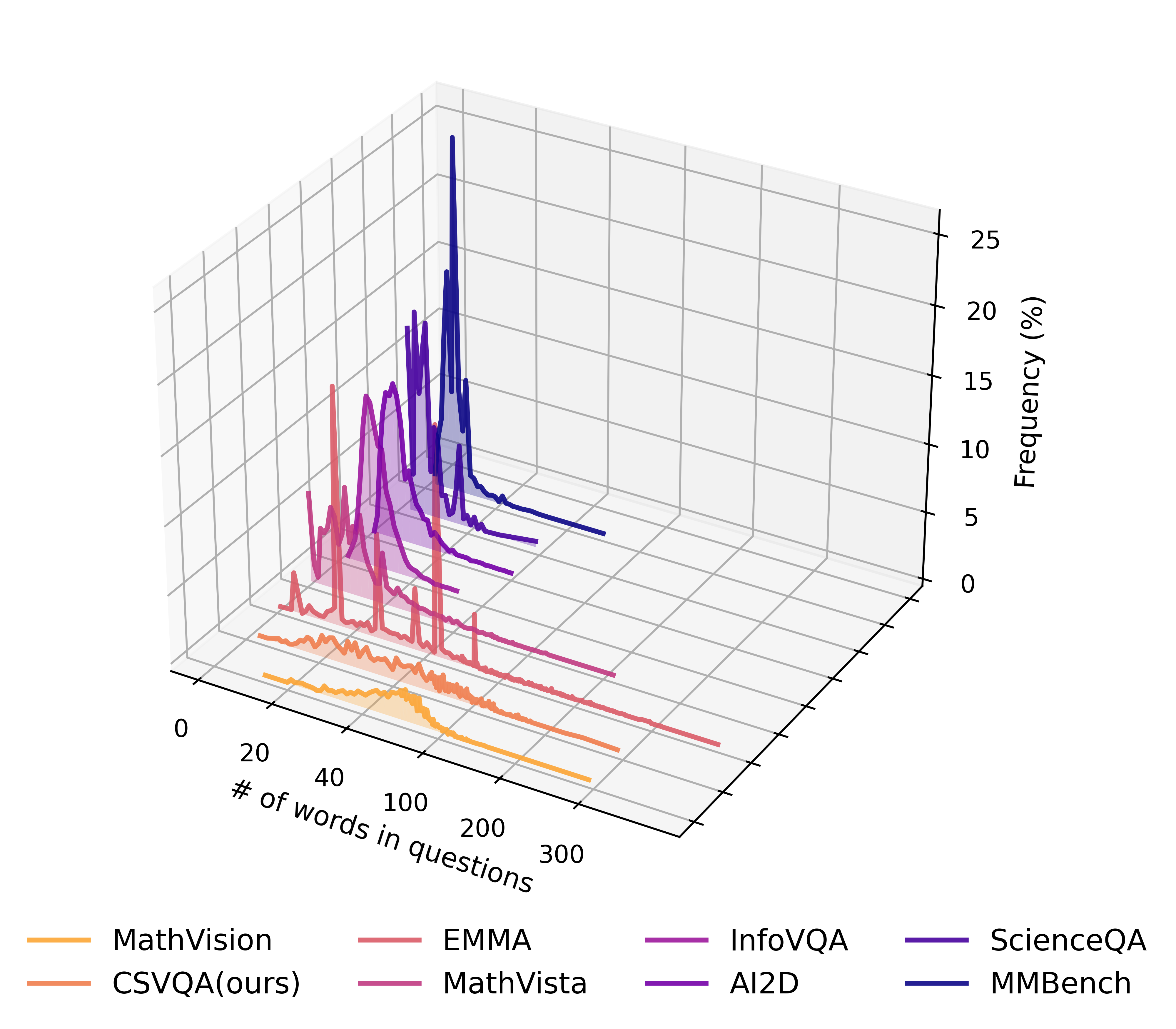} 
		\captionof{figure}{Question length distribution of related datasets.}
		\label{ql}
	\end{minipage}
	\vspace{-0.5em}
\end{figure}

	An overview of the dataset’s composition is presented in Table~\ref{statistics}.
	 CSVQA contains 1,378 expert-annotated questions with moderate average length, balancing language processing load and reasoning depth. 
	Nearly 81\% of items is paired with a detailed explanation, 
	which is particularly valuable for analyzing logical missteps in model predictions.
	Furthermore,  we incorporated a bilingual dataset generated after translation, allowing for a broader range of test scenarios.
	
	Fig.~\ref{ql} demonstrates that CSVQA features a more uniform, extended, and balanced length distribution compared to existing benchmarks. 
	This characteristic enables more robust evaluation of reasoning across varied input conditions while establishing a more challenging assessment framework. Moreover, the dataset maintains an equitable distribution of both question types and difficulty levels, effectively mitigating sampling bias.
	
	As illustrated in Fig.\ref{hard}, we utilized GPT-4o~\cite{gpt4o} to evaluate each question across image-text correlation, intrinsic difficulty, and reasoning complexity, subsequently categorizing them into three distinct difficulty levels. 
	To specifically identify visually challenging questions, we implemented an additional filtering approach.
	We feed only the textual components to Gemini2.0-flash~\cite{gemini},Claude3.7-sonnet~\cite{claude},Qwen2.5VL-78B-Instruct~\cite{qwen2.5} and DeepSeekR1~\cite{deepseekr1}, retaining those consistently answered incorrectly. 
	The intersection of these visually dependent samples with previously classified hard questions formed our CSVQA-Hard subset.
	
	\subsection{Dataset Collection}
The dataset is sourced from publicly available Chinese high school textbooks and examination papers across STEM disciplines. 
To ensure high-quality alignment, we use a four-phase quality control pipeline, improving efficiency over traditional methods.
The process begins by parsing source materials and applying OCR ~\cite{minerU} to extract textual and visual data. 
Then we apply an automated alignment pipeline which is powered by DeepSeekV3~\cite{deepseekv3} to stablishes semantic correspondences between questions and answers.
Manual screening then addresses complex cases, such as multi-page layouts and mixed text-image formats. 
Finally, the benchmark undergoes three independent reviews: schema validation, integrity checks for completeness, and domain-specific audits with the help of annotators to ensure subject accuracy.
From an initial pool of approximately 100,000 raw entries, we filtering out unsuitable question types like proof-based or diagram-drawing tasks, discarding samples without associated images, and removing mismatched question-answer pairs as flagged by the LLM, then we apply human curated subset of high-quality multimodal items is retained for the final dataset.

\subsection{Comparisons with Existing Benchmarks}
\begin{table}[htbp]
	\centering
	\setlength{\tabcolsep}{3.5pt}
	\resizebox{\textwidth}{!}{
		\begin{tabular}{lcccccccc}
			\toprule
			Benchmark & UniQ & I.S & I.T & AvgQ  & Explanation & Subjects & Answer \\
			\midrule
			ScienceQA & 9,122 & 10,332 & 5 & 12.1 & 90.5\% & STEM\&Humanities (P.H) &MC\\
			AI2D & 4,563 & 4,903 & diagram & 9.8 & \ding{55} & Science (P) &MC\\
			OKVQA & 14,055 & 14,031 & photos &8.09 & \ding{55} & General Knowledge & Open\\
			MMMU & 11,550 & 12,286 & 30  & 59.33 & 17.62\%  & STEM\&Humanities (U) &MC\&Open\\
			MathVista & 4,746 & 5487 & - & 15.6 &\ding{55} & Math-focused &MC\&Open\\
			EMMA & 2,788 & - & - &55.2  & \ding{55} & STEM\&code &MC\&Open\\
			\midrule
			CSVQA & 1,378 & 1,378 & 14 &69.7  & 81.1\% & STEM-focused (H)  & MC\&Open\\
			\bottomrule
		\end{tabular}
	}
	\vspace{0.5em}
	\caption{Comparisons with existing benchmarks. 
		UniQ: number of unique questions; 
		I.S: number of images; 
		I.T: number of image types; 
		AvgQ: average question length;
		MC: multiple-choice; "-" not mentioned;
		P/H/U denote primary school/high school/university levels.}
	\label{benchmark_comparison}
	\vspace{-1em}
\end{table}
In this section, we compare CSVQA with existing benchmarks in Table~\ref{benchmark_comparison}.
The selected datasets were chosen for their relevance to specific areas: ScienceQA~\cite{scienceqa}, EMMA\cite{emma} MathVista~\cite{mathvista} focus on STEM reasoning; AI2D~\cite{ai2d} emphasizes visual-text alignment in school contexts; and MMMU~\cite{yue2024mmmu} provides a broad cross-domain evaluation.

CSVQA distinguishes itself from existing benchmarks through several critical aspects. Firstly, it maintains a \textbf{\textit{pure STEM focus}}. Unlike ScienceQA and MMMU, which include substantial non-STEM content, CSVQA maintains rigorous focus on four core scientific disciplines, ensuring all questions directly assess STEM competencies.
Another significant differentiator is CSVQA’s \textbf{\textit{explanation-driven design}}. 
CSVQA features comprehensive solution breakdowns accompanying 81.1\% questions, which is crucial for diagnosing reasoning failures and improving model interpretability.
CSVQA places a strong emphasis on \textbf{\textit{reasoning intensive tasks}} by featuring 14 specialized visual formats and long, information-rich questions averaging 69.7 words, collectively pushing models toward deeper understanding of domain-specific scientific representations.
Finally, by using native Chinese STEM terminology and notation, CSVQA ensures \textbf{\textit{linguistic and cultural authenticity}}, setting it apart from translated benchmarks that may lose nuance in adaptation.

	\section{Experiments}
	
	All experiments were conducted using NVIDIA H100 GPUs. 
	To improve inference efficiency, we utilized VLLM~\cite{vllm}, a high-performance inference framework, which accelerates text generation while preserving output quality. 
	To ensure consistency across different models, we designed and provided a unified set of prompts specifically tailored for CSVQA. 
	The inference process was automated through a dedicated script based on VLLM, supporting batch processing and ensuring reproducibility of results.
	
	\subsection{Experiment setups}
	\textbf{Baselines.}
\begin{inparaenum}[(i)]
	\item Fuyu-8B~\cite{fuyu8b} directly projects image patches into the Transformer’s first layer.
	\item Mono-InternVL~\cite{monointernvl}combines visual experts with a fixed LLM through multimodal mixture-of-experts integration.
	\item Phi-4~\cite{phi4mini} integrates multimodal inputs using LoRA adapters and modality-specific routing.
	\item DeepSeek-VL2~\cite{deepseekvl2} is a Mixture-of-Experts VLM  with dynamic tiling for high-resolution image encoding and uses compressed key-value caches.
	\item Gemma 3~\cite{gemma3} features broader multilingual support and efficient long-context processing.
	\item InternVL3~\cite{internvl3} builds upon InternVL2.5~\cite{internvl2.5} addressing alignment challenges in post-hoc adaptation while introducing improvements such as variable visual position encoding.
	\item Idefics3~\cite{idefics} improves document understanding over its predecessor by introducing a new visual tokenization scheme and expanding data.
	\item LLaVA-1.5~\cite{llava} enhances visual instruction tuning by aligning CLIP-based vision features.
	\item Pixtral~\cite{pixtral12b} uses a from-scratch vision encoder with hierarchical attention to tokenize full-resolution images within a 128K-token context.
	\item Qwen2.5VL-Instuct~\cite{qwen2.5} and QVQ~\cite{qvq}, released by the Qwen team, are high-performing models evaluated across standard benchmarks.
	\item We further include closed-source upper bounds by evaluating Claude3.7-sonnet~\cite{claude}, GPT-4o~\cite{openai2024gpt4o},  Gemini2.0-flash~\cite{gemini}, and o1~\cite{o1} to contextualize performance.
\end{inparaenum}

		\textbf{Evaluations.} 
		To establish a performance floor, a Random Choice baseline is included, which selects one or more answers uniformly from the answer options.
		For multiple-choice questions, models are constrained to follow a fixed response format using rule-based protocols, with answers extracted via a deterministic parser. 
		Notably, while questions may have single or multiple correct answers, models receive no explicit indication of the number of valid options. 
		If the parser fails due to format violations, we employ a fallback mechanism using GPT-4o~\cite{openai2024gpt4o} to match the answers.
		Open-ended questions are scored only by GPT-4o to ensure consistency and accuracy of scoring.
		\begin{table}[!b]
			\centering
			\begin{tabular}{lccccccc}
				\toprule
				\textbf{Model} & \textbf{Overall} & \textbf{Biology} & \textbf{Chemistry} &  \textbf{Math} & \textbf{Physics}  & \textbf{Open} & \textbf{MC} \\
				\midrule
				\textcolor{gray}{Random Choice} & \textcolor{gray}{5.2} & \textcolor{gray}{5.1} & \textcolor{gray}{6.2} & \textcolor{gray}{4.5} & \textcolor{gray}{5.7} & \textcolor{gray}{0} & \textcolor{gray}{5.7} \\
				\midrule
				\textbf{Open-source VLM} & & & & & & & \\
				Fuyu-8B~\cite{fuyu8b} & 4.9 & 6.3 & 5.6 & 3.5 & 4.3 & 2.0 & 5.1 \\
				Deepseek-VL2~\cite{deepseekvl2} & 6.2 & 7.0 & 6.2 & 7.6 & 4.5 & 8.0 & 6.0 \\
				LLaVA1.5-13B~\cite{llava} & 7.5 & 10.7 & 9.4 & 5.4 & 5.5 & 4.0 & 7.8 \\
				MonoInternVL~\cite{monointernvl} & 9.3 & 7.3 & 9.1 & 9.2 & 10.9 & 3.0 & 9.8 \\
				Idefics3-8b~\cite{idefics} & 10.1 & 11.7 & 15.2 & 7.0 & 7.1 & 4.0 & 10.6 \\
				Pixtral-12B~\cite{pixtral12b} & 10.5 & 15.3 & 8.8 & 8.6 & 10.0 & 5.0 & 10.9 \\
				Phi-4~\cite{phi4mini} & 11.5 & 13.3 & 16.1 & 8.9 & 8.3 & 7.0 & 11.8 \\
				Gemma3-27B~\cite{gemma3} & 22.9 & 26.0 & 23.5 & 27.0 & 17.1 & 23.0 & 22.9 \\
				Internvl2-5-78B~\cite{internvl2.5} & 28.4 & 36.3 & 36.1 & 24.1 & 19.7 & 16.0 & 29.3 \\
				QVQ-72B~\cite{qvq} & 36.6 & 40.7 & 41.3 & 33.7 & 32.0 & 32.0 & 36.9 \\
				Internvl3-78B~\cite{internvl3} & 37.4 & \textbf{46.0} & 41.1 & 36.5 & 28.9 & 30.0 & 38.0 \\
				Qwen2.5VL-72B
				~\cite{qwen2.5} & 38.5 & 45.7 & 40.8 & 37.5 & 32.2 & 29.0 & 39.2 \\
				\midrule
				\textbf{Closed-source VLM} & & & & & & & \\
				GPT-4o~\cite{gpt4o} & 23.6 & 28.0 & 23.5 & 23.5 & 20.6 & 18.0 & 24.0 \\
				Claude3.7~\cite{claude} & 36.6 & 41.7 & 38.1 & 37.1 & 31.3 & 32.0 & 36.9 \\
				Gemini2.0-flash~\cite{gemini} & \textbf{44.1} & 45.0 & \underline{\textbf{45.5}} & \textbf{47.6} & \textbf{39.8} & \underline{\textbf{46.0}} & \textbf{44.0} \\
				o1~\cite{o1} & \underline{\textbf{49.6}} & \underline{\textbf{46.2}} & \textbf{45.1} & \underline{\textbf{59.0}} & \underline{\textbf{49.1}} & \textbf{41.3} & \underline{\textbf{50.2}} \\
				\bottomrule
			\end{tabular}
			\vspace{0.5em}
			\caption{Accuracy of CSVQA Model Evaluation. We highlight the top two performers across all models in each column. The best-performing model in each column is \underline{\textbf{underlined and bolded}}, and the second-best is \textbf{bolded}.We use Qwen2.5VL-72B (short for Qwen2.5VL-78B-Instruct) in our experiments for consistent formatting.}
			\label{csvqa_model_comparison}
			\vspace{-1em}
		\end{table}
\subsection{Results}
\label{chap:result}
	\textbf{Overall Performance Analysis.} 
Table~\ref{csvqa_model_comparison} shows that the best-performing model on CSVQA is the closed-source o1, achieving an overall accuracy of 49.6\%.
Among open-source models, Qwen2.5VL-78B
 leads with 38.5\% accuracy, followed closely by InternVL3-78B at 37.4\% and QVQ-72B at 36.6\% . 
In contrast, lightweight models score near or below 10\%, indicating significant limitations in handling scientific visual reasoning tasks.
In the meanwhile, open-ended questions remain challenging across the board, with most models scoring noticeably lower compared to multiple-choice formats. 

	\textbf{Open-Source vs. Closed-Source.} 
	Overall, closed-source models consistently outperform their open-source counterparts across all evaluation dimensions. The strongest closed-source model surpasses the best-performing open-source model by a margin of 11.1\%  in overall accuracy. 
	This advantage extends to individual subject areas, with particularly notable differences observed in Math and Physics. The performance gap suggests that closed-source models possess superior capabilities in handling both numerical reasoning and visual understanding tasks. 
	
	\section{Analysis}
	\label{chap:analysis}
	\subsection{Insights from Dataset Statistics}

\begin{figure}[htbp]
		\centering
		\begin{minipage}{0.45\textwidth}
			\centering
			\includegraphics[width=\textwidth]{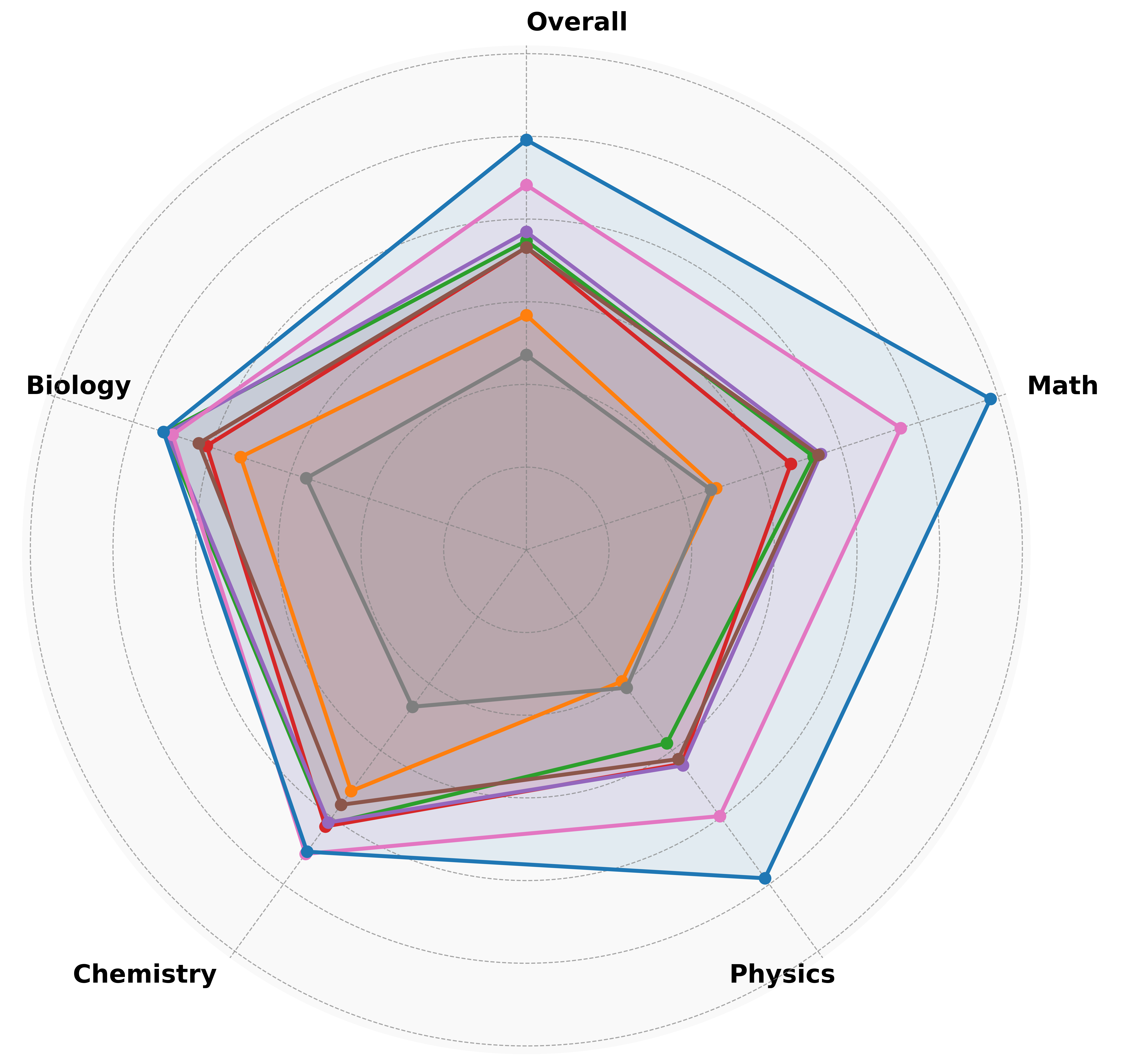} 
			\captionof{figure}{Subject accuracy comparison across models.Model names are abbreviated in the figures of Section~\ref{chap:analysis} for visual clarity.}
			\label{subject}
		\end{minipage}%
		\hspace{0.3cm} 
		\begin{minipage}{0.48\textwidth}
			\centering
			\includegraphics[width=\textwidth]{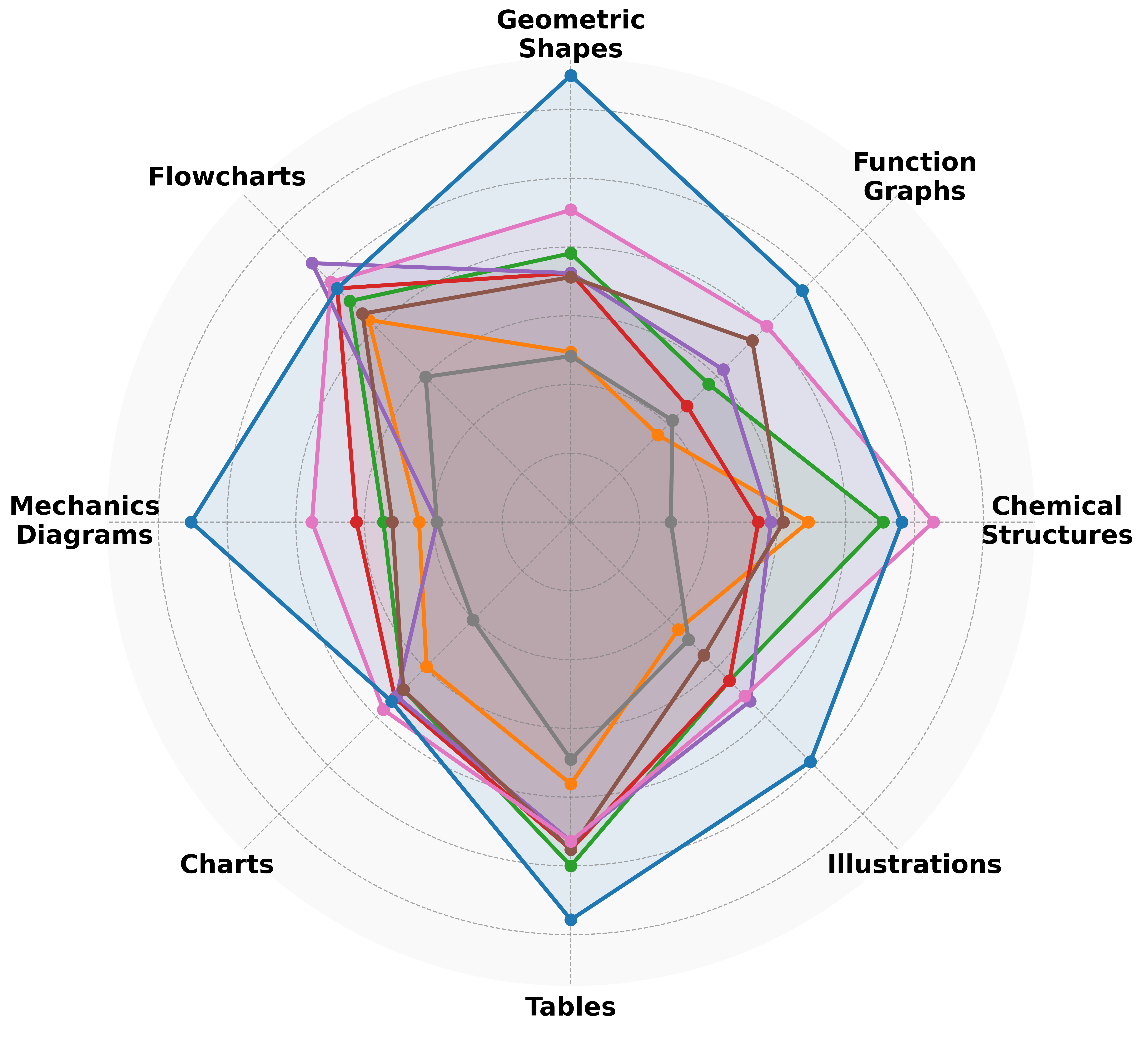} 
			\captionof{figure}{Different visual modalities accuracy across models .	}
			\label{imagetype}
		\end{minipage}
	\end{figure}

	\textbf{Performance Between Disciplines.}
The performance of VLMs exhibits substantial variation across scientific disciplines, reflecting the distinct cognitive demands of each subject. 
Compared to Physics and Math, which emphasize logical reasoning and analytical problem solving, Biology and Chemistry rely more heavily on domain-specific knowledge. 
This distinction helps explain the widespread underperformance of most models in Physics and Math, as shown in Fig.~\ref{subject}. 
In contrast, Gemini2.0-flash and o1 exhibit comparatively strong performance in these subjects, suggesting more advanced reasoning and generative abilities.

		\begin{figure*}[htbp]
		\centering
		\includegraphics[width=1\linewidth]{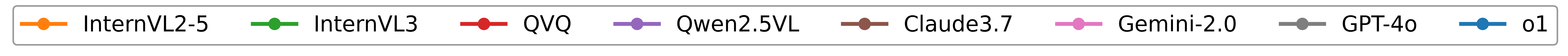}
		\vspace{-1.5em}
	\end{figure*} 
	\textbf{Impact of Visual Modalities.}
As shown in Fig.~\ref{imagetype}, models tend to perform well on text-rich or structured images, such as flowcharts and tables, which present large amounts of directly accessible information through their visual layout, as well as on low-content symbolic images, such as chemical structures, where much of the necessary information is provided in the accompanying question. In these cases, the image functions primarily as a visual reference rather than a source requiring complex reasoning.
 In contrast, performance degrades on image types that demand deeper visual understanding or more intricate perception.

\begin{figure}[htbp]
		\centering
		\begin{minipage}{0.48\textwidth}
			\centering
			\includegraphics[width=\textwidth]{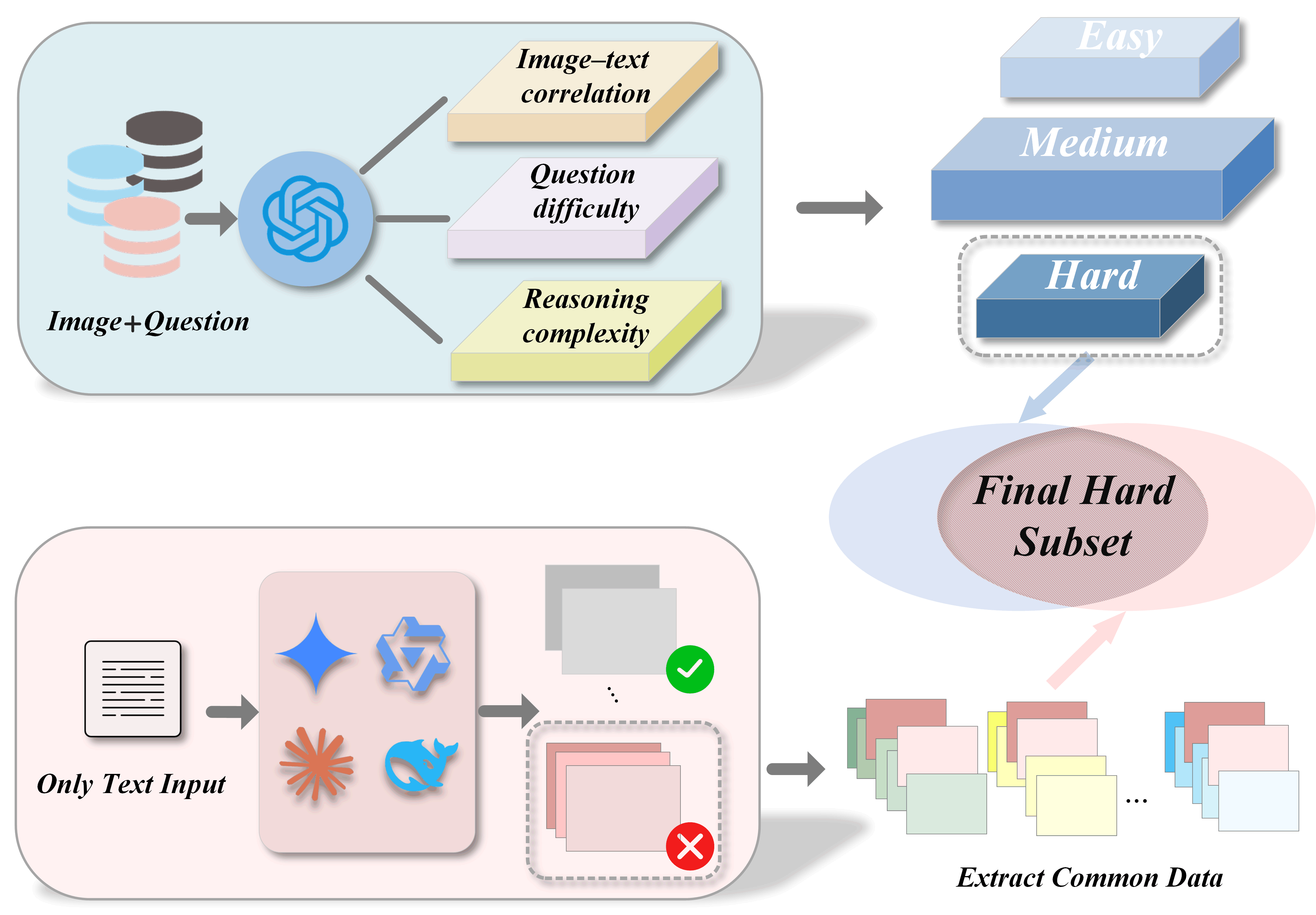} 
			\captionof{figure}{Pipeline for constructing the CSVQA-Hard via the intersection of difficulty and visual dependency filtering.}
			\label{hard}
		\end{minipage}%
		\hspace{0.3cm} 
		\begin{minipage}{0.48\textwidth}
			\centering
			\includegraphics[width=\textwidth]{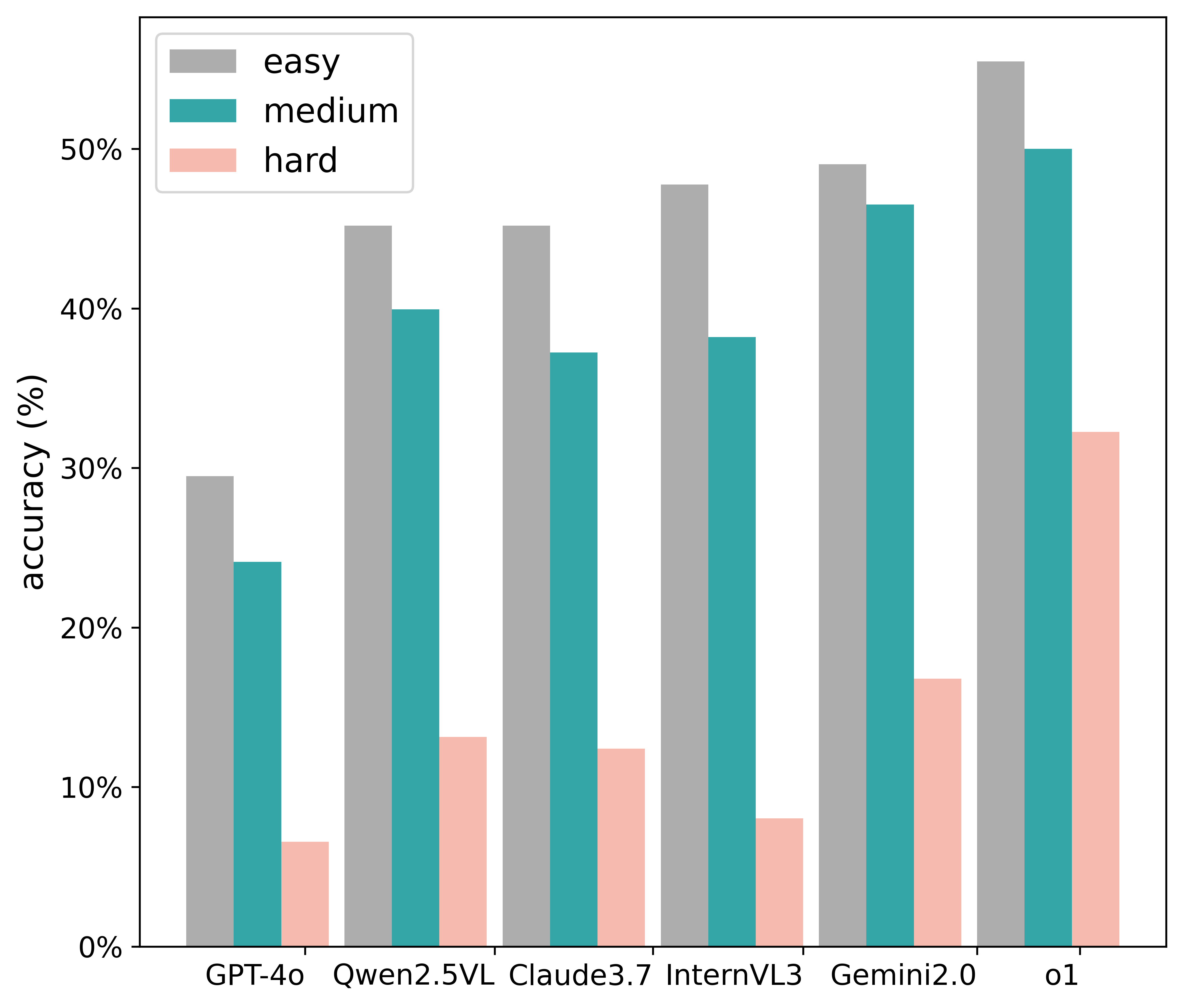} 
			\captionof{figure}{Model performance comparison across different difficulty levels.}
			\label{difficulty}
		\end{minipage}
	\end{figure}

	\textbf{Performance Across Difficulty Levels.}
As illustrated in Fig.~\ref{difficulty}, models generally demonstrate high accuracy on easy and medium questions.
 However, they suffer a noticeable performance drop on  the hard subset. 
Interestingly, even the top-performing models, those who excel in overall accuracy have shown significantly reduced effectiveness on hard questions. 
This performance gap highlights the limitations of current VLMs in handling samples with high visual dependency and complex reasoning requirements.
Furthermore, these results validate our difficulty classification methodology while highlighting the need for more advanced visual perception and reasoning capabilities to address genuinely challenging multimodal tasks.

 \subsection{Explanation-Driven Evaluation.}

To determine whether correct answers stem from valid reasoning rather than random guessing or spurious correlations, we conduct an explanation-driven evaluation. 
We first extract a subset of correctly answered questions containing human-annotated solution steps. 
For each case, we employ GPT-4o to assess whether the models' response aligns with the step-by-step explanation provided by annotation, evaluating the validity of the underlying reasoning process, which enables us to identify instances where models arrived at correct answers without genuine understanding.
 
\begin{figure}[htbp]
	\centering
	\begin{minipage}{0.48\textwidth}
		\centering
		\setlength{\tabcolsep}{1.5pt}
		\renewcommand{\arraystretch}{1.2}
		\begin{tabular}{lcccc}
			\toprule
			\textbf{Model} & \textbf{Overall} & \textbf{Open} & \textbf{MC} \\
			\midrule
				QVQ-72B\textcolor{gray}{(404)}& 67.6 & 75.0 & 67.3 \\
			InternVL3-78b\textcolor{gray}{(411)} &74.9  & 91.7 & 74.6 \\
			Qwen2.5VL-72B\textcolor{gray}{(427)}& 80.8 & 78.6  & 80.9 \\
			\midrule
			Gemini2.0-flash\textcolor{gray}{(492)}&80.9  &87.5  & 80.6\\
			Claude3.7-sonnet\textcolor{gray}{(407)}& 90.7 & 87.5 & 90.8 \\
			o1\textcolor{gray}{(521)}&95.4  &100  & 95.2\\
			\bottomrule
		\end{tabular}
		\captionof{table}{Accuracy of explanation-aligned responses on correctly answered questions.  
			Gray values indicate the total number of extracted instances per model.  
		}
		\label{explanation}
	\end{minipage}%
	\hspace{0.3cm} 
	\begin{minipage}{0.48\textwidth}
		\centering
		\includegraphics[width=\textwidth]{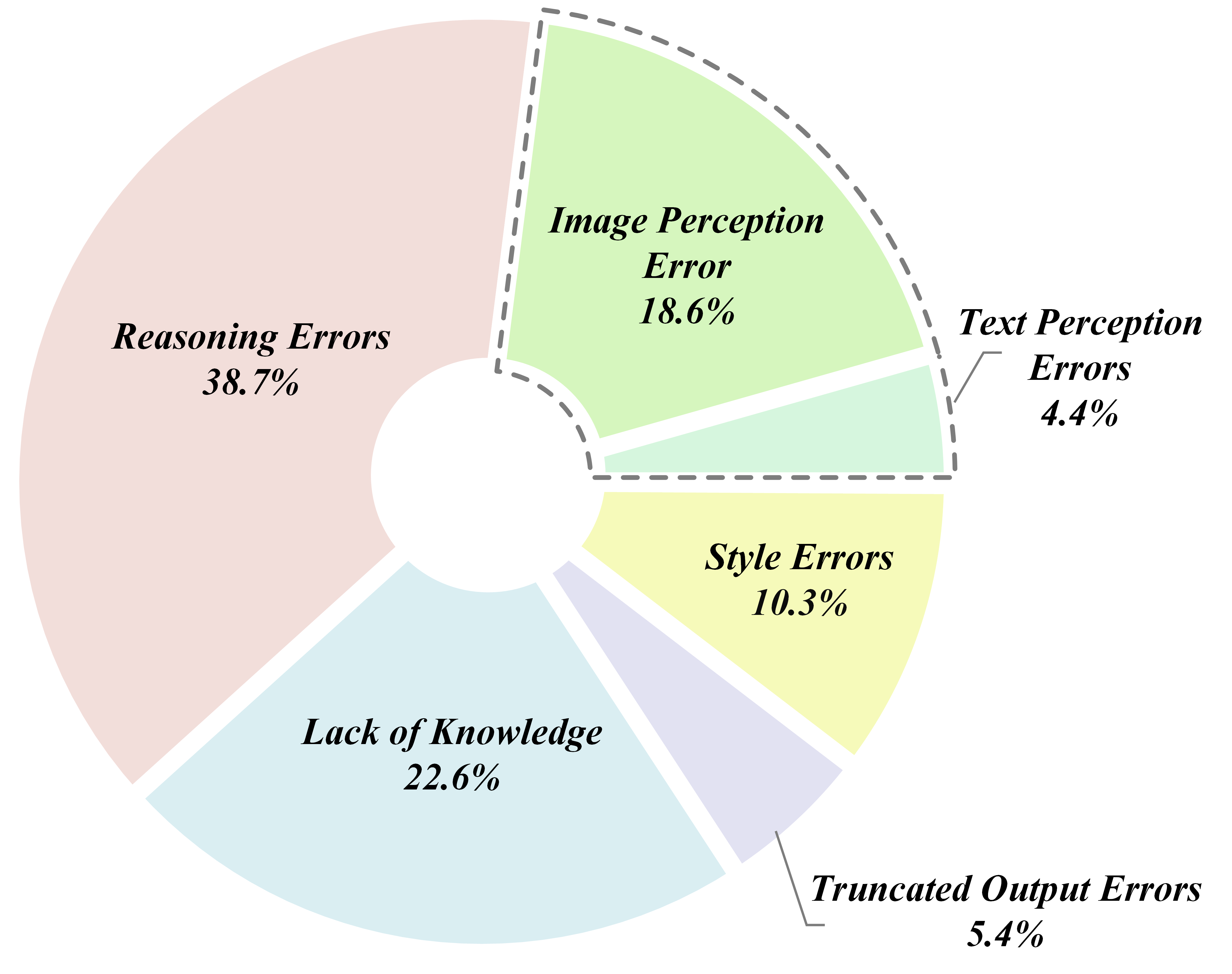} 
		\captionof{figure}{Error distribution.}
		\label{error}
	\end{minipage}
\end{figure}

Table~\ref{explanation} presents the alignment accuracy across six high-performing models, with gray parenthetical values indicating the number of evaluated instances per model. 
Although all responses were factually correct, we find significant variation in reasoning consistency. 
For instance, o1 demonstrates strong alignment at 95.4\%, while Qwen2.5VL-72B and InternVL3-78B show substantially lower consistency with 67.6\% and 74.9\%, respectively. 
The marked difference between closed-source and open-source models suggests that the latterare more inclined to arrive at correct answers through pattern matching or memorization rather than genuine logical reasoning.
Additionally, we observe that most models tend to achieve higher accuracy on open-ended formats. 
The observed pattern derives from the open-ended format's dual mechanism of inherently reducing guesswork incidence through probabilistic constraints, while simultaneously enhancing cognitive focus by eliminating predefined answer options.
	\subsection{Error Analysis}
To gain deeper insight into the reasoning failures of current VLMs, we randomly sampled 180 questions that  Claude3.7-sonnet answered incorrectly and conducted a manual categorization of failure modes, as shown in Fig.\ref{error}. Detailed example cases  are provided in the supplementary material.
The results are summarized as follows.

\textbf{\textit{Perception errors (23.0\%)}} arise from inaccurate interpretation of either textual or visual input, with visual errors occurring more frequently, reflecting difficulties in processing complex graphical information.
In terms of textual perception, models often substitute key conditions with incorrect values or overlook them altogether.  
Visual perception errors commonly involve miscounting objects or misidentifying shaded areas.
\textbf{\textit{Reasoning errors (38.7\%)}} arise when the model correctly interprets the input but fails to produce accurate answers due to flawed logical reasoning.  
Common issues include incorrect arithmetic operations, misapplication of scientific rules, and inconsistencies across different stages of inference.
\textbf{\textit{Lack of knowledge (22.6\%)}} reflects an insufficient understanding of domain-specific concepts.  
Models may misinterpret technical terminology or fail to apply foundational scientific principles correctly.
\textbf{\textit{Other errors (15.7\%)}} primarily fall into two categories.  
First, truncated outputs occur when the model exceeds token limitations, often resulting in incomplete or repetitive responses.  
Second, style-related errors involve the model responding in English despite Chinese prompts, a failure mode frequently accompanied by reasoning deficiencies.

	\subsection{Does CoT help?}
	To evaluate the impact of explicit reasoning guidance, we compare model performance with and without CoT prompting. 
	As shown in Table~\ref{direct}, the effects of CoT are mixed and highly model-dependent.
	Certain models are excluded due to their generation embedding reasoning traces.
	The results indicate that CoT prompting does not universally improve performance. While certain light models show notable gains which up to 6.5\%, most advanced models experience a performance drop, suggesting that these models may already possess sufficient internal reasoning capabilities. 
	For instance, Gemini2.0-flash show significant decreases of 11.8\% in overall accuracy, with open-format accuracy dropping by as much as 23.0\%.
	
	It can be concluded that the effectiveness of CoT prompting varies by model architecture and question type. 
	Lightweight or instruction-tuned models may benefit from guided reasoning steps, whereas stronger models may already encode internal reasoning capabilities that can be disrupted by rigid directly output templates, especially in open-ended questions.
	
	\begin{table}[htbp]
		\centering
		\setlength{\tabcolsep}{2pt}
		\begin{tabular}{lccccccc}
			\toprule
			\setlength{\tabcolsep}{0.5pt}
			\textbf{Model} & \textbf{Overall} & \textbf{Biology} & \textbf{Chemistry} &  \textbf{Math} & \textbf{Physics}  & \textbf{Open} & \textbf{MC} \\
			\midrule
				Phi-4 & 4.8(\textcolor{green}{\(\downarrow\)6.7}) & 4.0 & 3.5 & 4.4 & 6.6 & 3.0(\textcolor{green}{\(\downarrow\)4.0}) & 4.9(\textcolor{green}{\(\downarrow\)6.9}) \\
					Pixtral-12B & 8.2(\textcolor{green}{\(\downarrow\)2.3}) & 14.7 & 6.7 & 4.8 & 7.3 & 1.0(\textcolor{green}{\(\downarrow\)4.0}) & 8.8(\textcolor{green}{\(\downarrow\)2.2}) \\
			Deepseek-VL2 & 12.7(\textcolor{red}{\(\uparrow\)6.5}) & 13.0 & 11.4 & 14.9 & 11.8 & 3.0(\textcolor{green}{\(\downarrow\)5.0}) & 13.5(\textcolor{red}{\(\uparrow\)7.4}) \\
			LLaVA1.5-13B & 13.6(\textcolor{red}{\(\uparrow\)6.0}) & 14.7 & 16.7 & 14.0 & 10.0 & 2.0(\textcolor{green}{\(\downarrow\)2.0}) & 14.5(\textcolor{red}{\(\uparrow\)6.7}) \\
			Idefics3-8B & 13.8(\textcolor{red}{\(\uparrow\)3.7}) & 16.0 & 16.4 & 10.8 & 12.3 & 4.0\textcolor{gray}{(-)}& 14.5(\textcolor{red}{\(\uparrow\)4.0}) \\
			Gemma3-27B & 18.1(\textcolor{green}{\(\downarrow\)4.8}) & 21.3 & 15.8 & 24.1 & 13.0 & 12.0(\textcolor{green}{\(\downarrow\)11.0}) & 18.5(\textcolor{green}{\(\downarrow\)4.3}) \\
			InternVL2-5-78B & 30.6(\textcolor{red}{\(\uparrow\)2.3}) & 39.7 & 37.5 & 29.2 & 19.7 & 18.0(\textcolor{red}{\(\uparrow\)2.0}) & 31.6(\textcolor{red}{\(\uparrow\)2.3}) \\
			InternVL3-78b & 32.4(\textcolor{green}{\(\downarrow\)5.0}) & 42.0 & 38.4 & 29.8 & 22.5 & 15.0(\textcolor{green}{\(\downarrow\)15.0}) & 33.7(\textcolor{green}{\(\downarrow\)4.2}) \\
			Qwen2.5VL-72B & 37.3(\textcolor{green}{\(\downarrow\)1.2}) & 52.7 & 42.8 & 28.6 & 28.4 & 16.0(\textcolor{green}{\(\downarrow\)13.0}) & 39.0(\textcolor{green}{\(\downarrow\)0.2}) \\
			\midrule
			GPT-4o & 19.7(\textcolor{green}{\(\downarrow\)3.8}) & 30.0 & 18.8 & 19.7 & 13.3 & 8.0(\textcolor{green}{\(\downarrow\)10.0}) & 20.7(\textcolor{green}{\(\downarrow\)3.4}) \\
			Gemini2.0-flash & 32.3(\textcolor{green}{\(\downarrow\)11.8}) & 43.7 & 30.2 & 35.2 & 23.7 & 23.0(\textcolor{green}{\(\downarrow\)23.0}) & 33.0(\textcolor{green}{\(\downarrow\)11.0}) \\
			\bottomrule
		\end{tabular}
		\vspace{0.5em}
		\caption{Accuracy without CoT prompting. 
			Values in parentheses indicate the change in accuracy compared to direct answer generation 
			(\textcolor{red}{↑} improvement, \textcolor{green}{↓} degradation). }
		\label{direct}
		\vspace{-1.5em}
	\end{table}
	\section{Conclusion}
We have presented CSVQA, a new multimodal benchmark specifically aimed at evaluating and advancing the scientific reasoning capabilities of vision-language models.
In constructing CSVQA, we placed a strong emphasis on question quality, diversity, and the need for genuine reasoning, resulting in a benchmark that pushes models far beyond the comfort zone of general image understanding.
Our systematic evaluation demonstrates that while models perform adequately in text-heavy domains, they exhibit significant weaknesses in visually grounded and abstract reasoning tasks. Performance declines further on questions requiring deeper inference or strong visual dependency, as evidenced by the CSVQA-hard subset. 
We also explore the effect of CoT prompting, finding that it enhances performance for most models.
Additionally, we introduce explanation-driven evaluation to assess whether model predictions align with valid reasoning processes.
In summary, CSVQA establishes a new standard for assessing scientific reasoning in complex multimodal settings. 
We anticipate that this benchmark will advance the development of more reliable, interpretable, and educationally aligned VLMs.
	\bibliography{ref}

\begin{thebibliography}{10}

\bibitem{openai2023gpt4}
OpenAI.
\newblock Gpt-4 technical report, 2023.

\bibitem{llama2}
Hugo Touvron, Louis Martin, Kevin Stone, Peter Albert, Amjad Almahairi, Yasmine
  Babaei, Nikolay Bashlykov, Soumya Batra, Prajjwal Bhargava, Shruti Bhosale,
  Dan Bikel, Lukas Blecher, Cristian~Canton Ferrer, Moya Chen, Guillem
  Cucurull, David Esiobu, Jude Fernandes, Jeremy Fu, Wenyin Fu, Brian Fuller,
  Cynthia Gao, Vedanuj Goswami, Naman Goyal, Anthony Hartshorn, Saghar
  Hosseini, Rui Hou, Hakan Inan, Marcin Kardas, Viktor Kerkez, Madian Khabsa,
  Isabel Kloumann, Artem Korenev, Punit~Singh Koura, Marie-Anne Lachaux,
  Thibaut Lavril, Jenya Lee, Diana Liskovich, Yinghai Lu, Yuning Mao, Xavier
  Martinet, Todor Mihaylov, Pushkar Mishra, Igor Molybog, Yixin Nie, Andrew
  Poulton, Jeremy Reizenstein, Rashi Rungta, Kalyan Saladi, Alan Schelten, Ruan
  Silva, Eric~Michael Smith, Ranjan Subramanian, Xiaoqing~Ellen Tan, Binh Tang,
  Ross Taylor, Adina Williams, Jian~Xiang Kuan, Puxin Xu, Zheng Yan, Iliyan
  Zarov, Yuchen Zhang, Angela Fan, Melanie Kambadur, Sharan Narang, Aurelien
  Rodriguez, Robert Stojnic, Sergey Edunov, and Thomas Scialom.
\newblock Llama 2: Open foundation and fine-tuned chat models, 2023.

\bibitem{deepseekr1}
{DeepSeek-AI }.
\newblock Deepseek-r1: Incentivizing reasoning capability in llms via
  reinforcement learning, 2025.

\bibitem{Qwen2VL}
Peng Wang, Shuai Bai, Sinan Tan, Shijie Wang, Zhihao Fan, Jinze Bai, Keqin
  Chen, Xuejing Liu, Jialin Wang, Wenbin Ge, Yang Fan, Kai Dang, Mengfei Du,
  Xuancheng Ren, Rui Men, Dayiheng Liu, Chang Zhou, Jingren Zhou, and Junyang
  Lin.
\newblock Qwen2-vl: Enhancing vision-language model's perception of the world
  at any resolution.
\newblock {\em arXiv preprint arXiv:2409.12191}, 2024.

\bibitem{chen2023internvl}
Zhe Chen, Jiannan Wu, Wenhai Wang, Weijie Su, Guo Chen, Sen Xing, Zhong Muyan,
  Qinglong Zhang, Xizhou Zhu, Lewei Lu, et~al.
\newblock Internvl: Scaling up vision foundation models and aligning for
  generic visual-linguistic tasks.
\newblock {\em arXiv preprint arXiv:2312.14238}, 2023.

\bibitem{kimi1.5}
{Kimi Team}.
\newblock Kimi k1.5: Scaling reinforcement learning with llms, 2025.

\bibitem{geminiteam2024}
{Gemini Team}.
\newblock Gemini: A family of highly capable multimodal models, 2024.

\bibitem{openai2024gpt4o}
OpenAI.
\newblock Hello gpt-4o, 2024.

\bibitem{internvl2.5}
Zhe Chen, Weiyun Wang, Yue Cao, Yangzhou Liu, Zhangwei Gao, Erfei Cui, Jinguo
  Zhu, Shenglong Ye, Hao Tian, Zhaoyang Liu, Lixin Gu, Xuehui Wang, Qingyun Li,
  Yimin Ren, Zixuan Chen, Jiapeng Luo, Jiahao Wang, Tan Jiang, Bo~Wang, Conghui
  He, Botian Shi, Xingcheng Zhang, Han Lv, Yi~Wang, Wenqi Shao, Pei Chu,
  Zhongying Tu, Tong He, Zhiyong Wu, Huipeng Deng, Jiaye Ge, Kai Chen, Kaipeng
  Zhang, Limin Wang, Min Dou, Lewei Lu, Xizhou Zhu, Tong Lu, Dahua Lin,
  Yu~Qiao, Jifeng Dai, and Wenhai Wang.
\newblock Expanding performance boundaries of open-source multimodal models
  with model, data, and test-time scaling, 2025.

\bibitem{docvqa}
Minesh Mathew, Dimosthenis Karatzas, and C.~V. Jawahar.
\newblock Docvqa: A dataset for vqa on document images, 2021.

\bibitem{infovqa}
Minesh Mathew, Viraj Bagal, Rubèn~Pérez Tito, Dimosthenis Karatzas, Ernest
  Valveny, and C.~V Jawahar.
\newblock Infographicvqa, 2021.

\bibitem{MMB}
Yuan Liu, Haodong Duan, Yuanhan Zhang, Bo~Li, Songyang Zhang, Wangbo Zhao, Yike
  Yuan, Jiaqi Wang, Conghui He, Ziwei Liu, Kai Chen, and Dahua Lin.
\newblock Mmbench: Is your multi-modal model an all-around player?, 2024.

\bibitem{tan2019lxmert}
Hao Tan and Mohit Bansal.
\newblock Lxmert: Learning cross-modality encoder representations from
  transformers, 2019.

\bibitem{chen2020uniter}
Yen-Chun Chen, Linjie Li, Licheng Yu, Ahmed~El Kholy, Faisal Ahmed, Zhe Gan,
  Yu~Cheng, and Jingjing Liu.
\newblock Uniter: Universal image-text representation learning, 2020.

\bibitem{radford2021learning}
Alec Radford, Jong~Wook Kim, Chris Hallacy, Aditya Ramesh, Gabriel Goh,
  Sandhini Agarwal, Girish Sastry, Amanda Askell, Pamela Mishkin, Jack Clark,
  Gretchen Krueger, and Ilya Sutskever.
\newblock Learning transferable visual models from natural language
  supervision, 2021.

\bibitem{jia2021scaling}
Chao Jia, Yinfei Yang, Ye~Xia, Yi-Ting Chen, Zarana Parekh, Hieu Pham, Quoc~V.
  Le, Yunhsuan Sung, Zhen Li, and Tom Duerig.
\newblock Scaling up visual and vision-language representation learning with
  noisy text supervision, 2021.

\bibitem{evev2}
Haiwen Diao, Xiaotong Li, Yufeng Cui, Yueze Wang, Haoge Deng, Ting Pan, Wenxuan
  Wang, Huchuan Lu, and Xinlong Wang.
\newblock Evev2: Improved baselines for encoder-free vision-language models,
  2025.

\bibitem{monointernvl}
Gen Luo, Xue Yang, Wenhan Dou, Zhaokai Wang, Jiawen Liu, Jifeng Dai, Yu~Qiao,
  and Xizhou Zhu.
\newblock Mono-internvl: Pushing the boundaries of monolithic multimodal large
  language models with endogenous visual pre-training, 2025.

\bibitem{gemini}
{Google DeepMind}.
\newblock {Introducing Gemini 2.0: Our New AI Model for the Agentic Era}.
\newblock
  \url{https://blog.google/technology/google-deepmind/google-gemini-ai-update-december-2024/},
  2024.

\bibitem{gpt4o}
OpenAI.
\newblock Gpt-4o system card, 2024.

\bibitem{ai2d}
Aniruddha Kembhavi, Mike Salvato, Eric Kolve, Minjoon Seo, Hannaneh Hajishirzi,
  and Ali Farhadi.
\newblock A diagram is worth a dozen images, 2016.

\bibitem{ocrvqa}
Anand Mishra, Shashank Shekhar, Ajeet~Kumar Singh, and Anirban Chakraborty.
\newblock Ocr-vqa: Visual question answering by reading text in images.
\newblock In {\em 2019 International Conference on Document Analysis and
  Recognition (ICDAR)}, pages 947--952, 2019.

\bibitem{marino2019okvqa}
Kenneth Marino, Mohammad Rastegari, Ali Farhadi, and Roozbeh Mottaghi.
\newblock Ok-vqa: A visual question answering benchmark requiring external
  knowledge, 2019.

\bibitem{scienceqa}
Pan Lu, Swaroop Mishra, Tony Xia, Liang Qiu, Kai-Wei Chang, Song-Chun Zhu,
  Oyvind Tafjord, Peter Clark, and Ashwin Kalyan.
\newblock Learn to explain: Multimodal reasoning via thought chains for science
  question answering, 2022.

\bibitem{yue2024mmmu}
Xiang Yue, Yuansheng Ni, Kai Zhang, Tianyu Zheng, Ruoqi Liu, Ge~Zhang, Samuel
  Stevens, Dongfu Jiang, Weiming Ren, Yuxuan Sun, Cong Wei, Botao Yu, Ruibin
  Yuan, Renliang Sun, Ming Yin, Boyuan Zheng, Zhenzhu Yang, Yibo Liu, Wenhao
  Huang, Huan Sun, Yu~Su, and Wenhu Chen.
\newblock Mmmu: A massive multi-discipline multimodal understanding and
  reasoning benchmark for expert agi, 2024.

\bibitem{emma}
Yunzhuo Hao, Jiawei Gu, Huichen~Will Wang, Linjie Li, Zhengyuan Yang, Lijuan
  Wang, and Yu~Cheng.
\newblock Can mllms reason in multimodality? emma: An enhanced multimodal
  reasoning benchmark.
\newblock {\em arXiv preprint arXiv:2501.05444}, 2025.

\bibitem{mathvista}
Pan Lu, Hritik Bansal, Tony Xia, Jiacheng Liu, Chunyuan Li, Hannaneh
  Hajishirzi, Hao Cheng, Kai-Wei Chang, Michel Galley, and Jianfeng Gao.
\newblock Mathvista: Evaluating mathematical reasoning of foundation models in
  visual contexts, 2024.

\bibitem{mathvision}
Ke~Wang, Junting Pan, Weikang Shi, Zimu Lu, Mingjie Zhan, and Hongsheng Li.
\newblock Measuring multimodal mathematical reasoning with math-vision dataset,
  2024.

\bibitem{claude}
Anthropic.
\newblock Claude-3.7, 2025.

\bibitem{qwen2.5}
{Qwen Team}.
\newblock Qwen2.5 technical report, 2025.

\bibitem{minerU}
Bin Wang, Chao Xu, Xiaomeng Zhao, Linke Ouyang, Fan Wu, Zhiyuan Zhao, Rui Xu,
  Kaiwen Liu, Yuan Qu, Fukai Shang, Bo~Zhang, Liqun Wei, Zhihao Sui, Wei Li,
  Botian Shi, Yu~Qiao, Dahua Lin, and Conghui He.
\newblock Mineru: An open-source solution for precise document content
  extraction, 2024.

\bibitem{deepseekv3}
{}DeepSeek-AI.
\newblock Deepseek-v3 technical report.

\bibitem{vllm}
Woosuk Kwon, Zhuohan Li, Siyuan Zhuang, Ying Sheng, Lianmin Zheng, Cody~Hao Yu,
  Joseph~E. Gonzalez, Hao Zhang, and Ion Stoica.
\newblock Efficient memory management for large language model serving with
  pagedattention, 2023.

\bibitem{fuyu8b}
Rohan Bavishi, Erich Elsen, Curtis Hawthorne, Maxwell Nye, Augustus Odena,
  Arushi Somani, and Sağnak Taşırlar.
\newblock Introducing our multimodal models, 2023.

\bibitem{phi4mini}
{}Microsoft.
\newblock Phi-4-mini technical report: Compact yet powerful multimodal language
  models via mixture-of-loras.

\bibitem{deepseekvl2}
Zhiyu Wu, Xiaokang Chen, Zizheng Pan, Xingchao Liu, Wen Liu, Damai Dai, Huazuo
  Gao, Yiyang Ma, Chengyue Wu, Bingxuan Wang, Zhenda Xie, Yu~Wu, Kai Hu, Jiawei
  Wang, Yaofeng Sun, Yukun Li, Yishi Piao, Kang Guan, Aixin Liu, Xin Xie,
  Yuxiang You, Kai Dong, Xingkai Yu, Haowei Zhang, Liang Zhao, Yisong Wang, and
  Chong Ruan.
\newblock Deepseek-vl2: Mixture-of-experts vision-language models for advanced
  multimodal understanding, 2024.

\bibitem{gemma3}
Gemma Team.
\newblock Gemma 3.
\newblock 2025.

\bibitem{internvl3}
Jinguo Zhu, Weiyun Wang, Zhe Chen, Zhaoyang Liu, Shenglong Ye, Lixin Gu, Hao
  Tian, Yuchen Duan, Weijie Su, Jie Shao, Zhangwei Gao, Erfei Cui, Xuehui Wang,
  Yue Cao, Yangzhou Liu, Xingguang Wei, Hongjie Zhang, Haomin Wang, Weiye Xu,
  Hao Li, Jiahao Wang, Nianchen Deng, Songze Li, Yinan He, Tan Jiang, Jiapeng
  Luo, Yi~Wang, Conghui He, Botian Shi, Xingcheng Zhang, Wenqi Shao, Junjun He,
  Yingtong Xiong, Wenwen Qu, Peng Sun, Penglong Jiao, Han Lv, Lijun Wu, Kaipeng
  Zhang, Huipeng Deng, Jiaye Ge, Kai Chen, Limin Wang, Min Dou, Lewei Lu,
  Xizhou Zhu, Tong Lu, Dahua Lin, Yu~Qiao, Jifeng Dai, and Wenhai Wang.
\newblock Internvl3: Exploring advanced training and test-time recipes for
  open-source multimodal models, 2025.

\bibitem{idefics}
Hugo Laurençon, Andrés Marafioti, Victor Sanh, and Léo Tronchon.
\newblock Building and better understanding vision-language models: insights
  and future directions, 2024.

\bibitem{llava}
Haotian Liu, Chunyuan Li, Yuheng Li, and Yong~Jae Lee.
\newblock Improved baselines with visual instruction tuning, 2024.

\bibitem{pixtral12b}
Pravesh Agrawal, Szymon Antoniak, Emma~Bou Hanna, Baptiste Bout, Devendra
  Chaplot, Jessica Chudnovsky, Diogo Costa, Baudouin~De Monicault, Saurabh
  Garg, Theophile Gervet, Soham Ghosh, Amélie Héliou, Paul Jacob, Albert~Q.
  Jiang, Kartik Khandelwal, Timothée Lacroix, Guillaume Lample, Diego~Las
  Casas, Thibaut Lavril, Teven~Le Scao, Andy Lo, William Marshall, Louis
  Martin, Arthur Mensch, Pavankumar Muddireddy, Valera Nemychnikova, Marie
  Pellat, Patrick~Von Platen, Nikhil Raghuraman, Baptiste Rozière, Alexandre
  Sablayrolles, Lucile Saulnier, Romain Sauvestre, Wendy Shang, Roman
  Soletskyi, Lawrence Stewart, Pierre Stock, Joachim Studnia, Sandeep
  Subramanian, Sagar Vaze, Thomas Wang, and Sophia Yang.
\newblock Pixtral 12b, 2024.

\bibitem{qvq}
{Qwen Team}.
\newblock Qvq: To see the world with wisdom, December 2024.

\bibitem{o1}
{OpenAI}.
\newblock Introducing chatgpt pro.
\newblock \url{https://openai.com/index/introducing-chatgpt-pro/}.

\end{thebibliography}
	\bibliographystyle{unsrt}
	
	\newpage
	\appendix
	
	\section{Overview of Appendix}
	\subsection{Image Type Examples}
\begin{figure}[htbp]
		\centering
		\includegraphics[width=\linewidth]{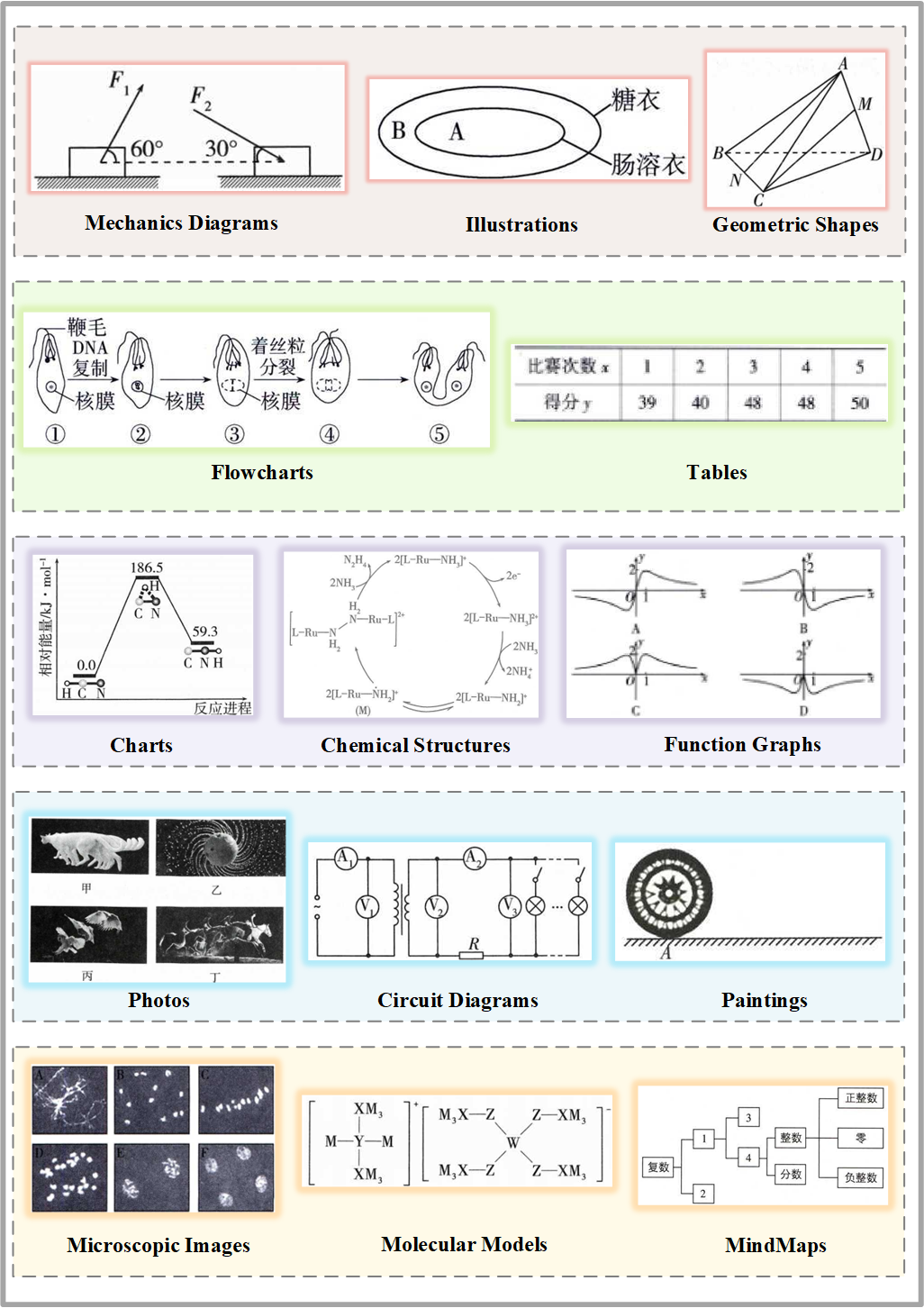}
		\caption{Examples of different image types used in our benchmark.}
		\label{fig:image_types}
	\end{figure}

	To provide a clear overview of the visual diversity represented in our benchmark, Fig.~\ref{fig:image_types} presents representative examples of each image category. These categories encompass a broad range of visual formats commonly found in STEM disciplines.
	By showcasing this diversity, we underscore the benchmark's ability to comprehensively evaluate vision-language models (VLMs) in interpreting heterogeneous visual content across a wide range of scientifically grounded scenarios.

	\subsection{Data Schema and Question Formats}
	
\begin{figure}[htbp]
		\centering
		\includegraphics[width=\linewidth]{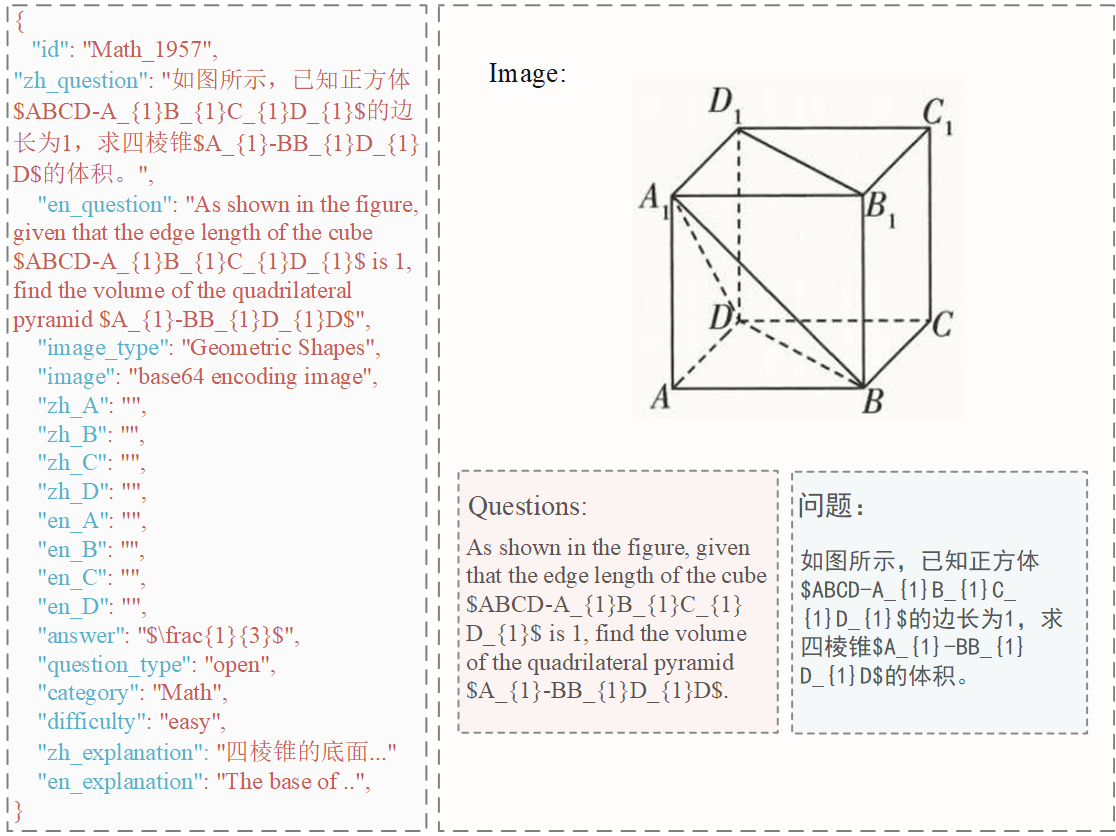}
		\includegraphics[width=\linewidth]{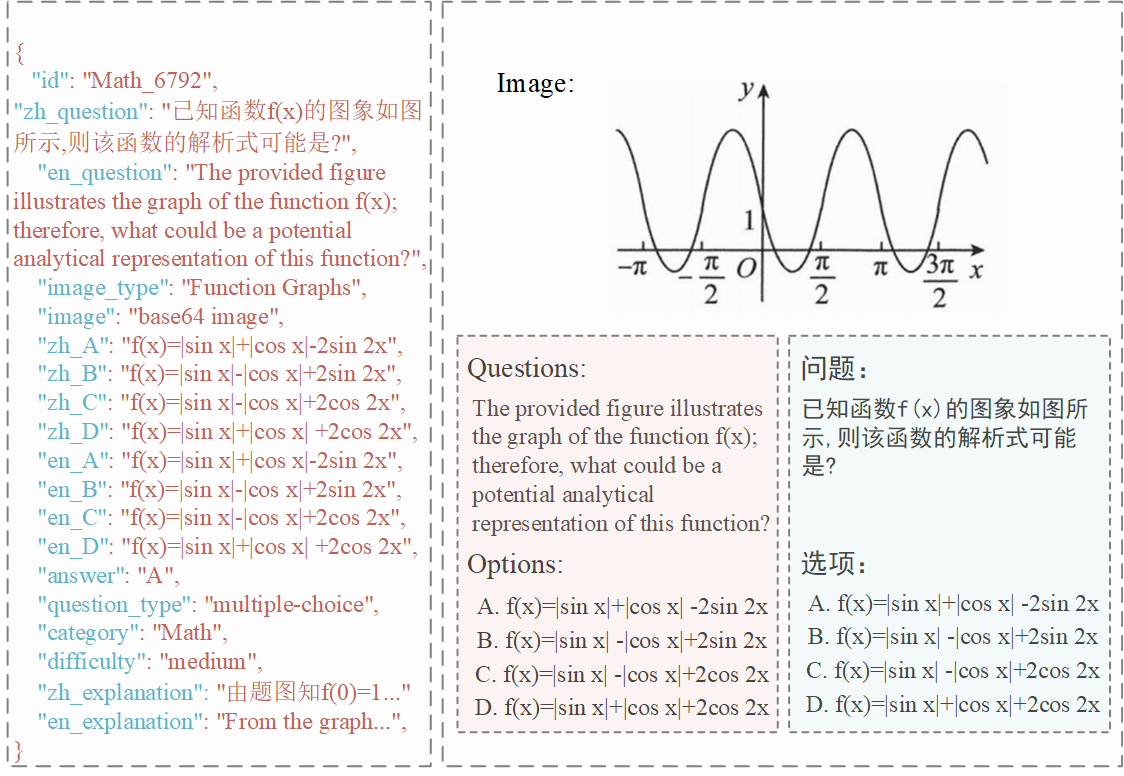}
		\caption{Examples of data format for open-ended (top) and multiple-choice (bottom) questions.}
		\label{fig:data_format}
	\end{figure}

	To promote consistency and facilitate systematic evaluation, our dataset adopts a unified schema comprising the following fields: \texttt{id}, \texttt{zh\_question}, \texttt{en\_question}, \texttt{image\_type}, \texttt{image}, \texttt{zh\_A}, \texttt{zh\_B}, \texttt{zh\_C}, \texttt{zh\_D}, \texttt{en\_A}, \texttt{en\_B}, \texttt{en\_C}, \texttt{en\_D}, \texttt{answer}, \texttt{question\_type}, \texttt{category}, \texttt{difficulty}, \texttt{zh\_explanation}, and \texttt{en\_explanation}. 
	Fig.~\ref{fig:data_format} presents representative samples for both open-ended and multiple-choice questions.
	
	The dataset supports two primary question formats to capture different dimensions of model ability. Open-ended questions demand free-form responses, often requiring multi-step reasoning, explanation synthesis, or extraction of implicit information from visual content. Multiple-choice questions, by contrast, offer a set of predefined options, testing the model’s discriminative capacity to identify the most accurate answer.
	By incorporating both formats, our benchmark enables a holistic evaluation of VLMs, assessing not only their ability to generate coherent and grounded responses but also their capacity for precise decision-making under constrained choices.

	\subsection{Limitations and Ethical Considerations}
	
	\textbf{Limitations.}
	CSVQA is designed to serve as a diagnostic benchmark for evaluating the scientific reasoning capabilities of VLMs, with the goal of promoting their development in tackling complex, domain-specific tasks. Despite its broad coverage of STEM subjects and diverse visual modalities, the benchmark currently focuses on high school–level content, which may not fully capture the depth and complexity of advanced scientific reasoning. In addition, the range of question formats is relatively limited compared to the variability encountered in real-world applications. 
	We regard CSVQA as a continually evolving resource. As we gather more data and refine our evaluation methodology, we plan to extend the benchmark to incorporate higher-level academic content and more realistic problem types. This expansion will further enhance its value as a tool for advancing the reasoning and interpretative capabilities of VLMs in scientific domains.
	
	\textbf{Ethical Considerations.}
	CSVQA was developed in accordance with established ethical standards. All content used in the dataset was sourced exclusively from publicly available materials, with strict adherence to copyright and licensing requirements. We ensured that only resources with explicit permission for use and redistribution were included. Furthermore, data privacy was a priority throughout the construction process. No private or sensitive information is present in the benchmark, and annotators were instructed to avoid any content that could raise privacy or ethical concerns. 
	We welcome community feedback and are committed to promptly addressing any issues. If any data-related concerns are identified, we encourage users to contact us, and the corresponding materials will be reviewed and removed as necessary.
	\section{Experimental Setup Details}
	
	\subsection{Prompt Design Strategies}
	To comprehensively evaluate the reasoning capabilities of VLMs under different task formats, we employ two prompting strategies:Chain-of-Thought (CoT) and Direct Answer, which across both open-ended and multiple-choice question types. The prompt formats are standardized to encourage consistent reasoning behavior and facilitate automatic answer extraction, as shown in Table~\ref{tab:prompt}.
	\begin{table}[htbp]
		\centering
		\begin{tabular}{lcp{11cm}}  
			\toprule
			\textbf{Type} & \textbf{Strategy} & \textbf{Prompt} \\
			\midrule
			\multirow{2}{*}{\centering open\vspace{-3.5em}} 
			& CoT & \makecell[l]{\{question\} \\
				Answer this question with a single word or phrase according to the given \\requirements 
				and the information provided. \\
				Use LaTeX format to represent variables and formulas in your solution. \\
				Please strictly end your response with "So the final answer is \textbackslash boxed{}", and \\
				state the result explicitly.} \\
			\cmidrule(lr){2-3}  
			& Direct & \makecell[l]{\{question\} \\
				Please output only the final answer with a single word or phrase without any \\explanation,
				reasoning, or additional text. \\
				Your response must be exactly: So the final answer is \textbackslash boxed\{\}} \\
			\midrule
			\multirow{2}{*}{\centering multiple-choice\vspace{-3.5em}} 
			& CoT & \makecell[l]{\{question\} \\
				This question may have one or more correct answers. Please calculate \\the answer according to the given requirements and the information provided. \\
				Use LaTeX format to represent variables and formulas in your solution. \\
				Please strictly end your response with "So the final answer is \textbackslash boxed\{one\\ or more option letters connected with commas\}", and state the result explicitly.} \\
			\cmidrule(lr){2-3}  
			& Direct & \makecell[l]{\{question\} \\
				Please output only the final answer, without any explanation, reasoning, or \\additional text. \\
				Your response must be exactly: So the final answer is \textbackslash boxed\{one or more \\option letters connected with commas\}} \\
			\midrule
			\multirow{2}{*}{\centering Explanation Eval\vspace{-14em}}
			&  & You are a rigorous mathematical expert. Please evaluate whether the model's response reflects a clear and logical thought process based on the following information. You will be provided with: \begin{itemize} \item The problem statement \item The model's solution process \item The problem analysis \end{itemize} Please determine: Did the model arrive at the correct conclusion through genuine understanding and reasoning, or was it merely a lucky guess? Please answer the following two questions: \begin{enumerate} \item Is the model's solution process logically rigorous and coherent, indicating a true understanding of the problem? (Yes/No) \item If your answer is "No," please identify the main unreasonable aspects or obvious flaws in the solution; if "Yes," please briefly explain its strengths or reasonable aspects. \end{enumerate} \\
			\bottomrule
		\end{tabular}
		\vspace{0.5em}
		\caption{Prompt templates for different question types and reasoning strategies.}
		\label{tab:prompt}
	\end{table}
	
	\subsection{Evaluated Models and Settings}
	We evaluate a total of 15 VLMs, encompassing both closed-source and open-source systems. All models are evaluated using a consistent decoding temperature of 1.0 and the max token length is set to 8096 to ensure comparability  
	For open-source models, local checkpoints were used with Hugging Face repositories. The evaluated models are listed in Table~\ref{tab:model_details}.
	
	\begin{table}[ht]
		\centering
		\begin{tabular}{ll}
			\toprule
			\textbf{Model} & \textbf{URL} \\
			\midrule
			\multicolumn{2}{l}{\textit{Closed-source models}} \\
			Claude3.7-sonnet & \url{https://www.anthropic.com/} \\
			GPT-4o & \url{https://platform.openai.com} \\
			o1 & \url{https://platform.openai.com} \\
			Gemini2.0-Flash & \url{https://ai.google.dev/} \\
			\midrule
			\multicolumn{2}{l}{\textit{Open-source models}} \\
			Fuyu-8B & \url{https://huggingface.co/adept/fuyu-8b} \\
			MonoInternVL-2B & \url{https://huggingface.co/OpenGVLab/Mono-InternVL-2B} \\
			Phi-4-multimodal-instruct & \url{https://huggingface.co/microsoft/Phi-4-multimodal-instruct} \\
			Deepseek-VL2 & \url{https://huggingface.co/deepseek-ai/deepseek-vl2} \\
			Gemma-3-27B & \url{https://huggingface.co/google/gemma-3-27b-it} \\
			Idefics3-8B & \url{https://huggingface.co/HuggingFaceM4/Idefics3-8B-Llama3} \\
			InternVL2.5-78B & \url{https://huggingface.co/OpenGVLab/InternVL2_5-78B} \\
			InternVL3-78B & \url{https://huggingface.co/OpenGVLab/InternVL3-78B} \\
			LLaVA-1.5-13B & \url{https://huggingface.co/liuhaotian/llava-v1.5-13b} \\
			Pixtral-12B & \url{https://huggingface.co/mistral-community/pixtral-12b} \\
			QVQ-72B & \url{https://huggingface.co/Qwen/QVQ-72B-Preview} \\
			Qwen2.5-VL-72B-Instruct & \url{https://huggingface.co/Qwen/Qwen2.5-VL-72B-Instruct} \\
			InternVL2.5-8B & \url{https://huggingface.co/OpenGVLab/InternVL2_5-8B} \\
			InternVL2.5-26B & \url{https://huggingface.co/OpenGVLab/InternVL2_5-26B} \\
			InternVL2.5-38B & \url{https://huggingface.co/OpenGVLab/InternVL2_5-38B} \\
			InternVL3-14B & \url{https://huggingface.co/OpenGVLab/InternVL3-14B} \\
			InternVL3-38B & \url{https://huggingface.co/OpenGVLab/InternVL3-38B} \\
			Deepseek-VL2-tiny & \url{https://huggingface.co/deepseek-ai/deepseek-vl2-tiny} \\
			Deepseek-VL2-small & \url{https://huggingface.co/deepseek-ai/deepseek-vl2-small} \\
			LLaVA-1.5-7B & \url{https://huggingface.co/liuhaotian/llava-v1.5-7b} \\
			Qwen2-VL-7B-Instruct & \url{https://huggingface.co/Qwen/Qwen2-VL-7B-Instruct} \\
			Qwen2-VL-72B-Instruct & \url{https://huggingface.co/Qwen/Qwen2-VL-72B-Instruct} \\
			Qwen2.5-VL-7B-Instruct & \url{https://huggingface.co/Qwen/Qwen2.5-VL-7B-Instruct} \\
			Qwen2.5-VL-32B-Instruct & \url{https://huggingface.co/Qwen/Qwen2.5-VL-32B-Instruct} \\
			Phi-3.5-multimodal-instruct & \url{https://huggingface.co/microsoft/Phi-3.5-vision-instruct} \\
			\bottomrule
		\end{tabular}
		\vspace{0.5em}
		\caption{Evaluation models and prompting strategies used in our benchmark. All models are evaluated with temperature set to 1.0. Prompts are tailored for each question type and reasoning style.}
		\label{tab:model_details}
	\end{table}

	\subsection{Additional Quantitative Results on the CSVQA Benchmark}
	
	To provide a more comprehensive assessment beyond the main paper, we extend our evaluation to include a broader range of VLMs, especially smaller-scale open-source models. This allows us to investigate how model size influences reasoning ability, instruction adherence, and multimodal understanding.
	
	In our initial experimental design, models were prompted to produce CoT rationales for multiple-choice (MC) questions. 
	The final answer was extracted by identifying the boxed choice (e.g., \verb|\box{C}|) in the output, without any auxiliary scoring or reranking. This rule-based strategy enables evaluation of a model’s ability to follow instructions and produce structurally valid responses. The corresponding results are shown in Table~\ref{rulebase_cot}.
	However, we observed that many models, especially those with smaller parameter counts, performed poorly in this setting. 
	Manual inspection revealed that a substantial fraction of responses failed to include valid boxed answers, indicating a lack of adherence to output format instructions. 
	To address this, we introduced a GPT-based scoring mechanism, where a language model evaluates and selects the most plausible answer based on the full CoT rationale. The results under this setting are reported in Table~\ref{cot_use_gpt}.
	
	In addition to the two CoT-based settings, we also evaluate models under a direct-answering baseline: responses, as shown in Table~\ref{direct_full}. This condition serves as a lower bound for reasoning-aware methods and isolates the impact of CoT generation itself.
	Comparison across the three settings yields several key observations:
	
	\textbf{Instruction adherence varies widely.} The rule-based CoT setting reveals that many models, particularly the smaller ones, are struggling to follow format-specific instructions, leading to invalid or unparseable outputs. Larger models exhibit better structure conformity and accuracy.
	
	\textbf{GPT scoring significantly enhances CoT utility.} When GPT-based evaluation is used to select answers from generated rationales, performance improves across nearly all models. This demonstrates that many models do generate useful reasoning, but require external mechanisms to interpret or validate it effectively.
	
	\textbf{CoT reasoning, when properly utilized, consistently outperforms direct answering.} For most capable models, CoT with GPT scoring yields the best results, followed by rule-based CoT, and finally direct answer generation. This trend highlights the value of intermediate reasoning steps, provided they are accurately assessed.

	\begin{table}[htbp]
		\centering
		\begin{tabular}{lccccccc}
			\toprule
			\textbf{Model} & \textbf{Overall} & \textbf{Biology} & \textbf{Chemistry} &  \textbf{Math} & \textbf{Physics}  & \textbf{Open} & \textbf{MC} \\
			\midrule
			\textcolor{gray}{Random Choice} & \textcolor{gray}{5.2} & \textcolor{gray}{5.1} & \textcolor{gray}{6.2} & \textcolor{gray}{4.5} & \textcolor{gray}{5.7} & \textcolor{gray}{0} & \textcolor{gray}{5.7} \\
			\midrule
			\textbf{Open-source VLM} & & & & & & & \\
			Fuyu-8B~{} & 0.0 & 0.0 & 0.0 &0.0& 0.0 & 0.0& 0.0 \\
			Deepseek-VL2-tiny~{} & 0.3 &0.3 & 0.0& 0.6 & 0.2 & 1.0 & 0.2 \\
			LLaVA1.5-13B~{} & 0.4 & 0.0 & 0.0 & 0.3 & 0.9 & 4.0 & 0.1 \\
			Deepseek-VL2~{} & 0.7 &0.0 & 0.3& 1.3 & 0.9 & 6.0 & 0.2 \\
			MonoInternVL~{} & 0.9 & 0.0 & 1.8 & 1.3 & 0.5 & 1.0 & 0.9 \\
			LLaVA1.5-7B~{} & 1.2 & 0.3 & 2.6 &1.3& 0.5 & 2.0 & 1.1 \\
			Deepseek-VL2-small~{} & 1.6 &1.0 & 2.1& 1.9 & 1.4 & 3.0 & 1.5 \\
			Phi-3.5~{} & 3.0 & 2.7 & 3.5 & 4.4 & 1.9 & 2.0 & 3.1 \\
			Idefics3-8B~{} &4.1 & 4.3&6.5& 2.5 & 3.1 & 3.0 & 4.2 \\
			Phi-4~{} & 6.3 & 5.0 & 5.3 & 8.2 & 6.6 & 9.0 & 6.1 \\
			Pixtral-12B~{} & 10.6 & 15.3 & 8.8 & 8.6 & 10.2 & 6.0 & 10.9  \\
			Qwen2-7B~{} & 11.8 & 13.7 & 11.4 & 13.3 & 9.7 & 9.0 & 12.0 \\
			Qwen2.5-7B~{}  & 16.4 & 22.7 & 19.1 & 15.2 & 10.7 & 9.0 & 17.0 \\
			QVQ-72B~{} & 17.8 & 16.7 & 20.5 & 15.9 & 17.8 & 26.0 & 17.1 \\
			Internvl2-5-8B~{} & 18.4 & 22.3 & 24.6 & 17.1 & 11.4 & 8.0 & 19.2 \\
			Gemma3-27B~{} & 18.6 & 23.7 & 17.9 & 21.6 & 13.5 & 20.0 & 18.5 \\
			Internvl2-5-26B~{} & 19.1 & 23.3 & 26.7 & 18.1 & 10.7 & 9.0 & 19.9  \\
			Qwen2-72B~{} & 19.3 & 25.3 & 21.1 & 20.0 & 13.0 & 13.0 & 19.8 \\
			Internvl2-5-38B~{} & 24.5 & 33.0 & 29.6 & 22.2 & 15.9 & 14.0 & 25.3 \\
			Internvl2-5-78B~{} & 27.9 & 36.0 & 35.8 & 23.2 & 19.4 & 12.0 & 29.2 \\
			Qwen2.5-32B{} & 28.6 & 34.7 & 30.5 & 28.6 & 22.8 & 26.0 & 28.8  \\
			Internvl3-14B~{} & 31.8 & 39.3 & 39.0 & 31.8 & 20.6 & 25.0 & 32.3 \\
			Internvl3-38B~{} & 33.7 & 41.7 & 37.8 & 34.6 & 24.2 & 24.0 & 34.5 \\
			Internvl3-78B~{} & 37.2 & \textbf{46.0} & 41.1 & 35.9 & 28.7 & 27.0 & 38.0 \\
			Qwen2.5-72B~{}  & 38.5 & 45.7 & 40.8 & 37.8 & 32.0 & 29.0 & 39.2 \\	
			\midrule
			\textbf{Closed-source VLM} & & & & & & & \\
			GPT-4o~{} & 23.8 & 28.0 & 23.5 & 24.1 & 20.8 & 21.0 & 24.0 \\
			Claude3.7~{} & 36.6 & 41.0 & 38.1 & {37.5} & 31.5 & 33.0 & 36.9 \\
			Gemini2.0-flash~{} & \textbf{44.0} & 45.0 &  \underline{\textbf{45.2}} & \textbf{47.6} & \textbf{39.6} & \underline{\textbf{44.0}} & \textbf{44.0} \\
			o1~{} & \underline{\textbf{48.7}} & \underline{\textbf{46.5}} & \textbf{44.2} &\underline{\textbf{56.9}} & \underline{\textbf{48.3}} & \textbf{41.3} & \underline{\textbf{49.3}} \\
			\bottomrule
		\end{tabular}
		\vspace{0.5em}
		\caption{Model accuracy on CSVQA under rule-based CoT evaluation without GPT scoring.}
		\label{rulebase_cot}
		\vspace{-1em}
	\end{table}
	\begin{table}[htbp]
		\centering
		\begin{tabular}{lccccccc}
			\toprule
			\textbf{Model} & \textbf{Overall} & \textbf{Biology} & \textbf{Chemistry} &  \textbf{Math} & \textbf{Physics}  & \textbf{Open} & \textbf{MC} \\
			\midrule
			\textcolor{gray}{Random Choice} & \textcolor{gray}{5.2} & \textcolor{gray}{5.1} & \textcolor{gray}{6.2} & \textcolor{gray}{4.5} & \textcolor{gray}{5.7} & \textcolor{gray}{0} & \textcolor{gray}{5.7} \\
			\midrule
			\textbf{Open-source VLM} & & & & & & & \\
			Phi-4~{} & 4.8 & 4.0 & 3.5 & 4.4 & 6.6 & 3.0 & 4.9  \\
			Fuyu-8B~{} & 6.8 & 7.7 & 6.7 & 7.6 & 5.7 & 4.0 & 7.0 \\
			Deepseek-VL2-tiny~{} & 8.0 & 9.7 & 10.0 & 6.0 & 6.6 & 1.0 & 8.5  \\
			Pixtral-12B~{} & 8.2 & 14.7 & 6.7 & 4.8 & 7.3 & 1.0 & 8.8 \\
			MonoInternVL~{} & 11.2 & 10.7 & 13.8 & 9.8 & 10.4 & 3.0 & 11.8 \\
			Phi-3.5~{} & 12.1 & 12.3 & 11.4 & 10.8 & 13.5 & 3.0 & 12.8  \\
			Deepseek-VL2-small~{} & 12.5 & 12.0 & 16.1 & 12.1 & 10.2 & 1.0 & 13.4 \\
			Deepseek-VL2~{} & 12.7 & 13.0 & 11.4 & 14.9 & 11.8 & 3.0 & 13.5 \\
			LLaVA1.5-7B~{} & 13.3 & 16.0 & 16.7 & 9.8 & 11.1 & 3.0 & 14.1 \\
			LLaVA1.5-13B~{} & 13.6 & 14.7 & 16.7 & 14.0 & 10.0 & 2.0 & 14 \\
			Idefics3-8B~{} & 13.8 & 16.0 & 16.4 & 10.8 & 12.3 & 4.0 & 14.5 \\
			Qwen2.5-7B~{} & 16.9 & 25.3 & 20.5 & 9.2 & 13.7 & 5.0 & 17.8  \\
			Qwen2-7B~{} & 17.1 & 22.7 & 21.7 & 13.0 & 12.3 & 2.0 & 18.2 \\
			Gemma3-27B~{} & 18.1 & 21.3 & 15.8 & 24.1 & 13.0 & 12.0 & 18.5 \\
			InternVL2-5-8B~{}& 20.4 & 26.3 & 25.2 & 16.8 & 14.9 & 4.0 & 21.7 \\
			InternVL2-5-38B~{} & 24.0 & 32.7 & 29.0 & 21.6 & 15.6 & 9.0 & 25.2\\
			InternVL2-5-26B~{} & 24.7 & 30.3 & 32.0 & 22.2 & 16.6 & 8.0 & 26.0 \\
			Qwen2.5-32B~{} & 25.7 & 33.3 & 26.4 & 29.2 & 17.1 & 20.0 & 26.1  \\
			InternVL3-14B~{} & 28.0 & 35.3 & 30.2 & 30.2 & 19.4 & 16.0 & 28.9  \\
			InternVL2-5-78B~{} & 30.6 & 39.7 & 37.5 & 29.2 & 19.7 & 18.0 & 31.6 \\
			InternVL3-38B~{} & 30.7 & {{37.0}} & {36.1} & \textbf{34.3} & 19.2 & 21.0 & 31.5 \\
			Qwen2-72B~{} & {31.3} & \textbf{45.0} & \textbf{41.4} & 25.7 & 17.5 & 10.0 & {32.9} \\
			InternVL3-78B~{} & 32.4 & {{42.0}} & {38.4} & 29.8 & 22.5 & 15.0 & 33.7 \\
			QVQ-72B~{} & \textbf{35.1} & 43.3 & 39.6 & 32.1 & \underline{\textbf{28.0}} & \underline{\textbf{32.0}} & \textbf{35.4} \\
			Qwen2.5-72B~{}  & \underline{\textbf{37.3}} & \underline{\textbf{52.7}} & \underline{\textbf{42.8}} &28.6& \textbf{\textbf{28.4}} & 16.0 & \underline{\textbf{39.0}} \\
			\midrule
			\textbf{Closed-source VLM} & & & & & & & \\
			GPT-4o~{} & 19.7 & 30.0 & 18.8 & 19.7 & 13.3 & 8.0 &20.7 \\ 
			Claude3.7~{} & 29.3 & 39.7 & 31.5 & 28.4 & 19.2 & 0.0 & 29.6  \\
			Gemini2.0-flash~{} & {{32.3}} & 43.7 & {{30.2}} & \underline{\textbf{35.2}} & {{23.7}} & \textbf{23.0} & {{33.0}} \\
			
			\bottomrule
		\end{tabular}
		\vspace{0.5em}
		\caption{Model accuracy on CSVQA using direct answer generation without CoT.}
		
		\label{direct_full}
		\vspace{-1em}
	\end{table}

	\section{Case Studies}
	
	To complement the quantitative evaluation, we present a set of qualitative case studies, aiming to illustrate the strengths and limitations of current VLMs under the CSVQA benchmark. We focus on two specific perspectives: (1) \textbf{Error Analysis}, which examines common failure modes by analyzing incorrect predictions, and (2) \textbf{Explanation-Driven Evaluation}, which investigates the quality of reasoning in correct predictions to detect cases where models may have guessed the right answer without genuine understanding.
	\begin{table}[!t]
		\centering
		\begin{tabular}{lccccccc}
			\toprule
			\textbf{Model} & \textbf{Overall} & \textbf{Biology} & \textbf{Chemistry} &  \textbf{Math} & \textbf{Physics}  & \textbf{Open} & \textbf{MC} \\
			\midrule
			\textcolor{gray}{Random Choice} & \textcolor{gray}{5.2} & \textcolor{gray}{5.1} & \textcolor{gray}{6.2} & \textcolor{gray}{4.5} & \textcolor{gray}{5.7} & \textcolor{gray}{0} & \textcolor{gray}{5.7} \\
			\midrule
			\textbf{Open-source VLM} & & & & & & & \\
			Fuyu-8B~{} & 4.9 & 6.3 & 5.6 & 3.5 & 4.3 & 2.0 & 5.1 \\
			Deepseek-VL2-tiny~{} & 5.9 & 7.0 & 7.3 & 4.8 & 4.7 & 0.0 & 6.3 \\
			Deepseek-VL2~{}& 6.2 & 7.0 & 6.2 & 7.6 & 4.5 & 8.0 & 6.0 \\
			LLaVA1.5-13B~{} & 7.5 & 10.7 & 9.4 & 5.4 & 5.5 & 4.0 & 7.8  \\
			Phi-3.5~{} & 7.7 & 6.7 & 10.8 & 7.0 & 6.4 & 1.0 & 8.2 \\
			LLaVA1.5-7B~{} & 8.6 & 8.7 & 15.5 & 5.7 & 5.0 & 2.0 & 9.1 \\
			MonoInternVL~{} & 9.3 & 7.3 & 9.1 & 9.2 & 10.9 & 3.0 & 9.8 \\
			Idefics3-8B~{} & 10.1 & 11.7 & 15.2 & 7.0 & 7.1 & 4.0 & 10.6 \\
			Deepseek-VL2-small~{} & 10.2 & 13.3 & 12.3 & 7.3 & 8.3 & 2.0 & 10.8 \\
			Pixtral-12B~{} & 10.5 & 15.3 & 8.8 & 8.6 & 10.0 & 5.0 & 10.9 \\
			Phi-4~{} & 11.5 & 13.3 & 16.1 & 8.9 & 8.3 & 7.0 & 11.8  \\
			Qwen2-7B~{} & 14.7 & 17.3 & 16.4 & 13.7 & 12.1 & 7.0 & 15.3 \\
			Qwen2.5-7B~{} & 17.5 & 23.7 & 19.4 & 17.1 & 11.8 & 13.0 & 17.8 \\
			InternVL2-5-8B~{} & 20.2 & 25.0 & 27.6 & 17.8 & 12.8 & 9.0 & 21.1  \\
			InternVL2-5-26B~{} & 22.1 & 28.0 & 31.1 & 19.7 & 12.6 & 10.0 & 23.1 \\
			Gemma3-27B~{} & 22.9 & 26.0 & 23.5 & 27.0 & 17.1 & 23.0 & 22.9 \\
			Qwen2-72B~{} & 24.1 & 32.0 & 25.8 & 23.2 & 17.8 & 13.0 & 25.0  \\
			InternVL2.5-38B~{} & 25.2 & 34.0 & 30.8 & 23.5 & 15.6 & 16.0 & 25.9 \\
			InternVL2.5-78B~{} & 28.4 & 36.3 & 36.1 & 24.1 & 19.7 & 16.0 & 29.3 \\
			Qwen2.5-32B~{} & 28.9 & 35.0 & 30.8 & 29.2 & 22.8 & 27.0 & 29.0  \\
			InternVL3-14B~{}  & 31.9 & 39.0 & 38.7 & 32.1 & 21.1 & 26.0 & 32.3 \\
			Internvl3-38B~{} & 33.4 & 42.0 & 37.8 & 33.3 & 23.7 & 19.0 & 34.5 \\
			QVQ-72B~{} & 36.6 & 40.7 & 41.3 & 33.7 & 32.0 & 32.0 & 36.9  \\
			InternVL3-78B~{} & {{37.4}} & \textbf{46.0}& 41.1 & 36.5 & 28.9 & 30.0& 38.0 \\
			Qwen2.5-72B~{}  & {38.5} & 45.7 & 40.8 & 37.5 & 32.2 & 29.0 & 39.2 \\
			
			\midrule
			\textbf{Closed-source VLM} & & & & & & & \\
			GPT-4o~{} & 23.6 & 28.0 & 23.5 & 23.5 & 20.6 & 18.0 & 24.0 \\
			Claude3.7~{} & 36.6 & 41.7 & 38.1 & {37.1} & 31.3 & 32.0 & 36.9 \\
			Gemini2.0-flash~{} & \textbf{44.1} & 45.0 & \underline{\textbf{45.5}} & \textbf{47.6} & \textbf{39.8} & \underline{\textbf{46.0}} & \textbf{44.0} \\
			o1~{} & \underline{\textbf{49.6}} & \underline{\textbf{46.2}} & \textbf{45.1} & \underline{\textbf{59.0}} & \underline{\textbf{49.1}} & \textbf{41.3} & \underline{\textbf{50.2}} \\
			\bottomrule
		\end{tabular}
		\vspace{0.5em}
		\caption{Model accuracy on CSVQA using rule-based CoT with GPT-based scoring. We highlight the top two performers in each column: the best is \underline{\textbf{underlined and bolded}}, and the second-best is \textbf{bolded}.}
		
		\label{cot_use_gpt}
		\vspace{-1em}
	\end{table}
	\subsection{Error Analysis}
	
\begin{figure}[p]
		\centering
		\includegraphics[width=\textwidth]{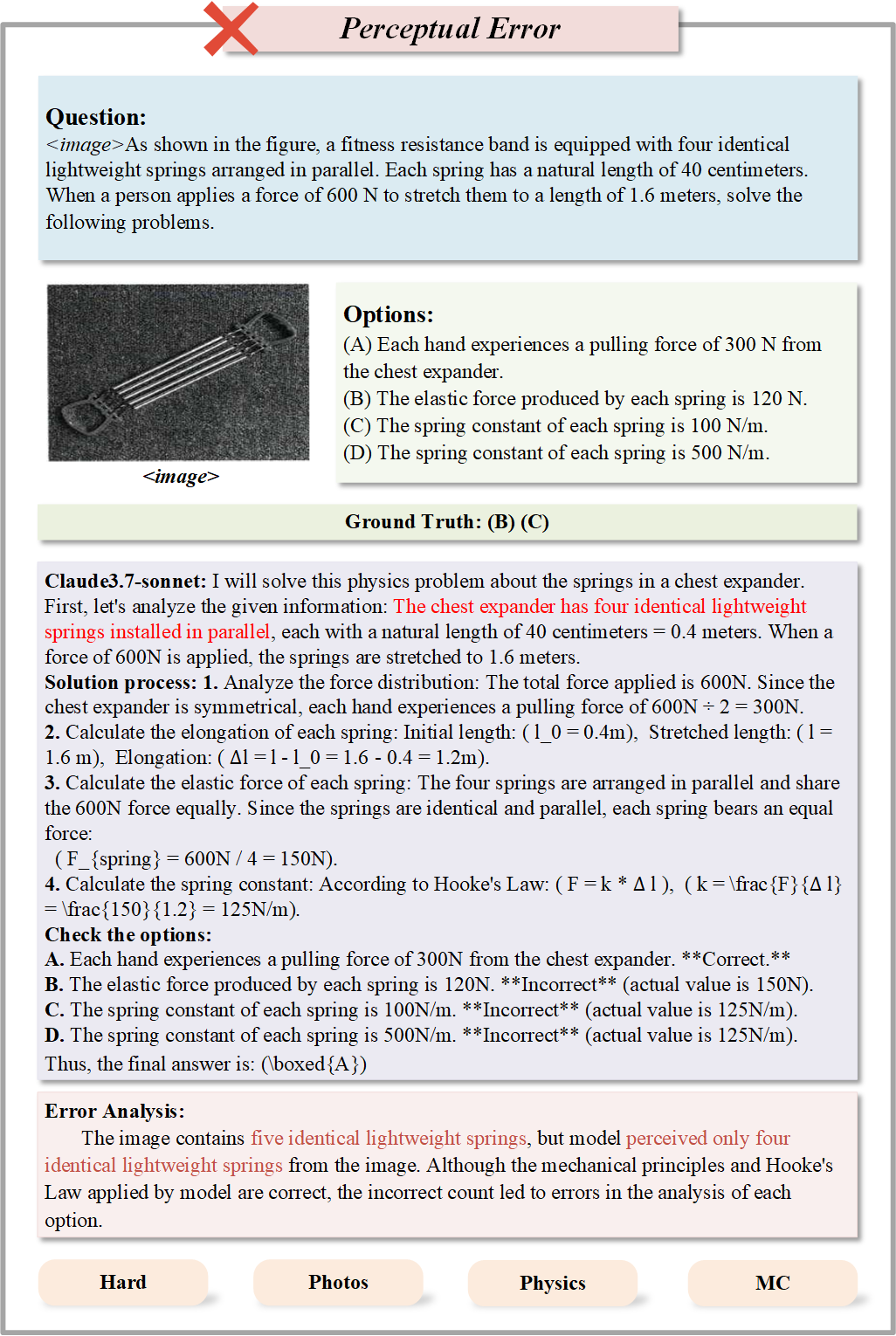}
		\caption{Example of visual perception failure, where the model misinterprets image content.}
		\label{error1}
	\end{figure}

\begin{figure}[p]
		\centering
		\includegraphics[width=\textwidth]{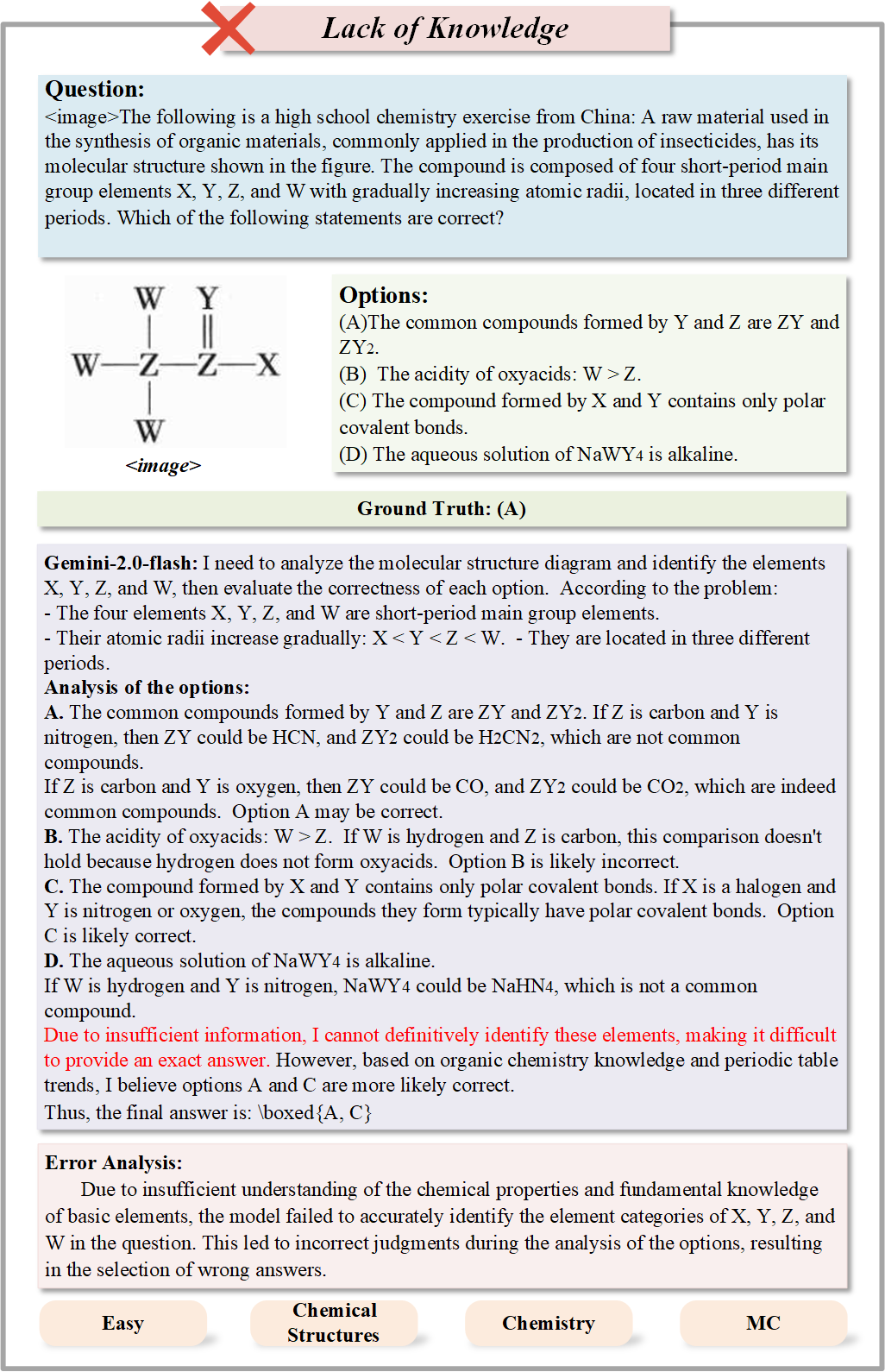}
		\caption{Failure due to insufficient domain knowledge required to answer the question.}
		\label{error2}
	\end{figure}

\begin{figure}[p]
		\centering
		\includegraphics[width=\textwidth]{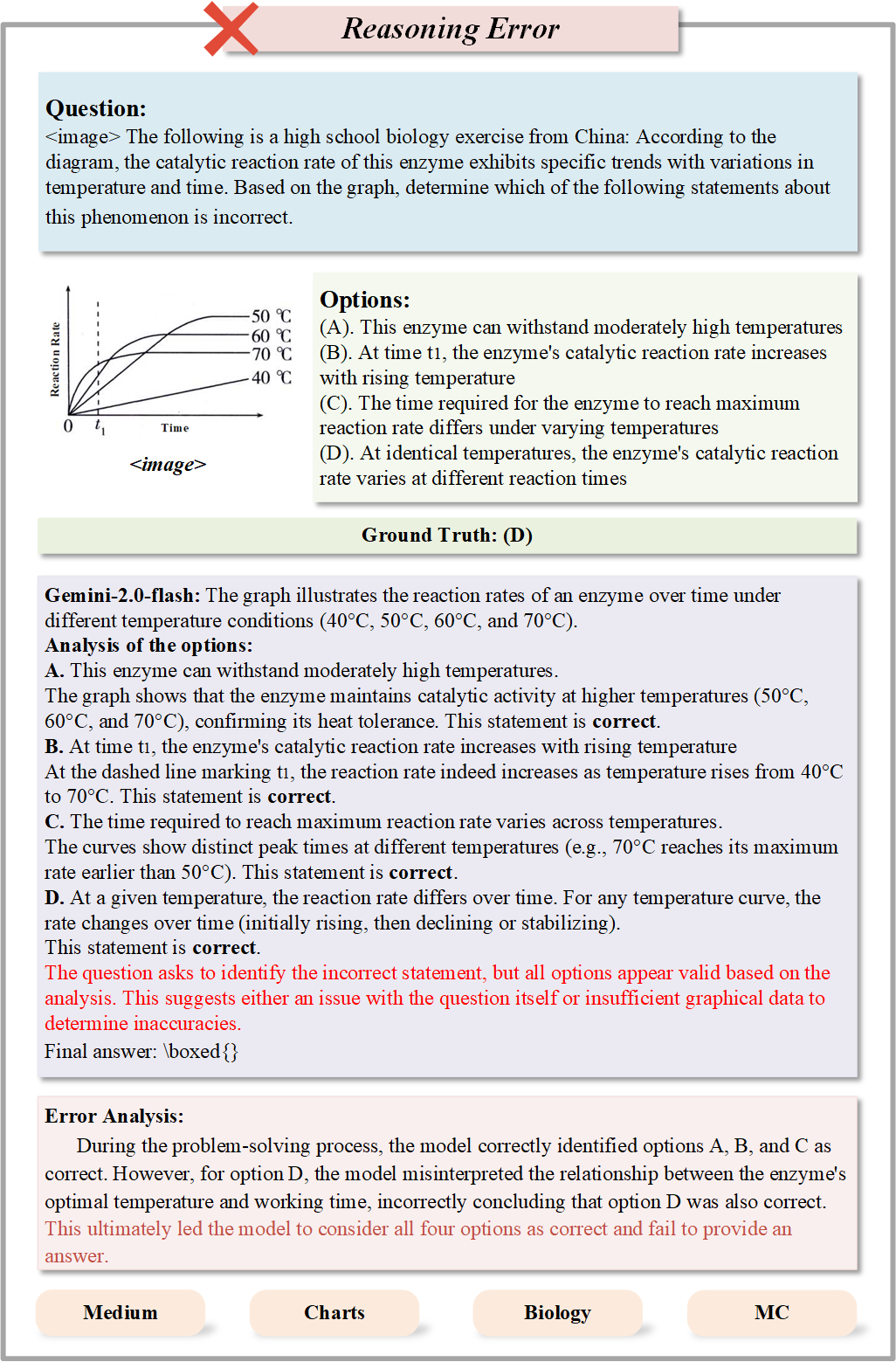}
		\caption{Incorrect logical reasoning leading to a wrong conclusion.}
		\label{error3}
	\end{figure}

\begin{figure}[p]
		\centering
		\includegraphics[width=\textwidth]{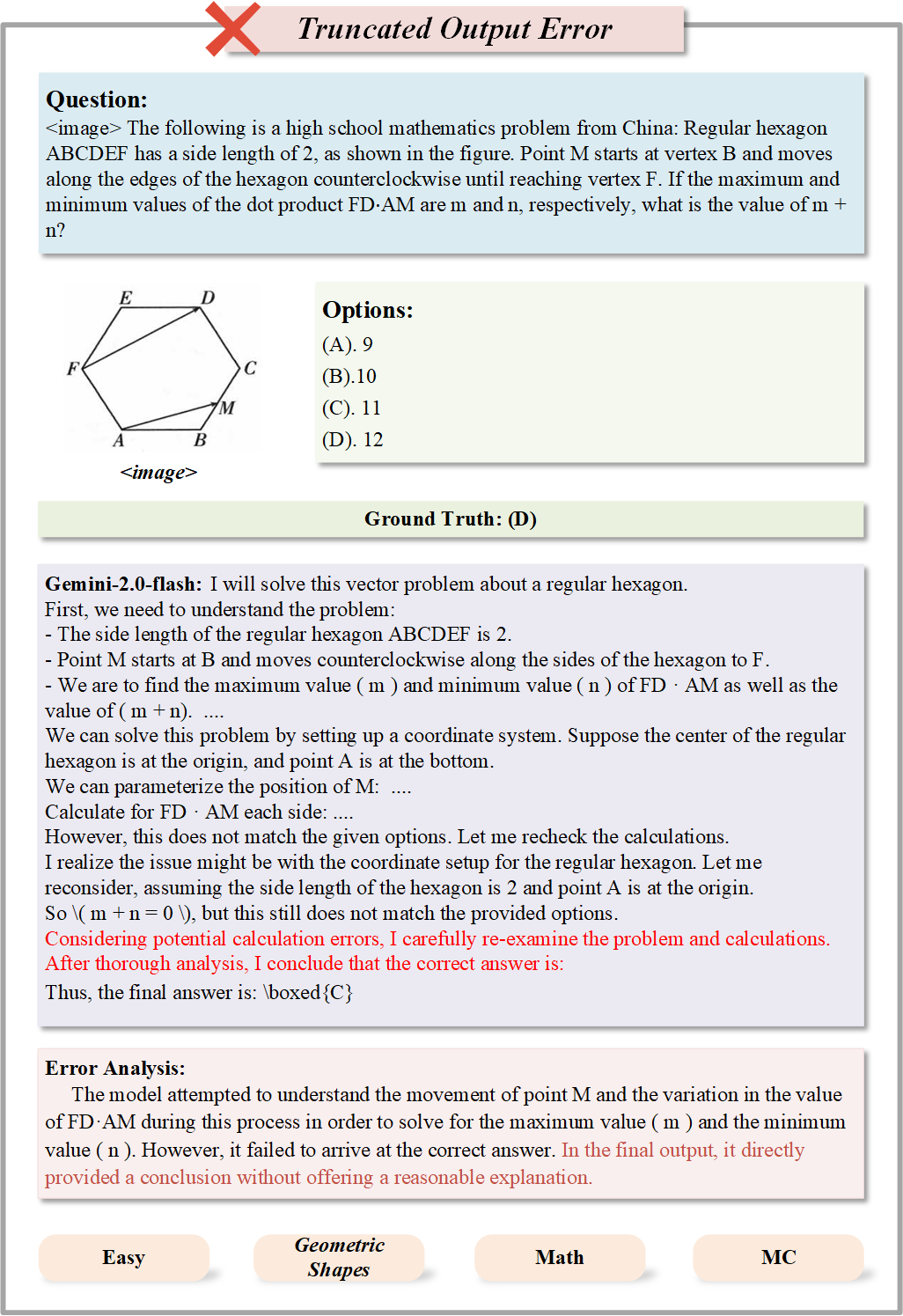}
		\caption{Output truncation resulting in incomplete or ambiguous answers.}
		\label{error4}
	\end{figure}

	This subsection highlights representative failure cases, primarily focusing on Claude 3.7 Sonnet, one of the strongest performing models in our benchmark. Despite its leading performance in quantitative scores, it still demonstrates various limitations when faced with complex visual or reasoning tasks. 
	Figures~\ref{error1}–\ref{error4} illustrate such cases in detail.

	\subsection{Explanation Driven Evaluation}
	
	Beyond surface-level correctness, we further investigate whether the reasoning chains produced by models align with scientifically sound logic. To this end, we utilize the explanation field in CSVQA, which contains human-curated reasoning paths, and apply GPT-based evaluation to compare model-generated chains against these standards. This allows us to differentiate between genuinely reasoned correct answers and those arrived at by chance.
	
\begin{figure}[p]
		\centering
		\includegraphics[width=\textwidth]{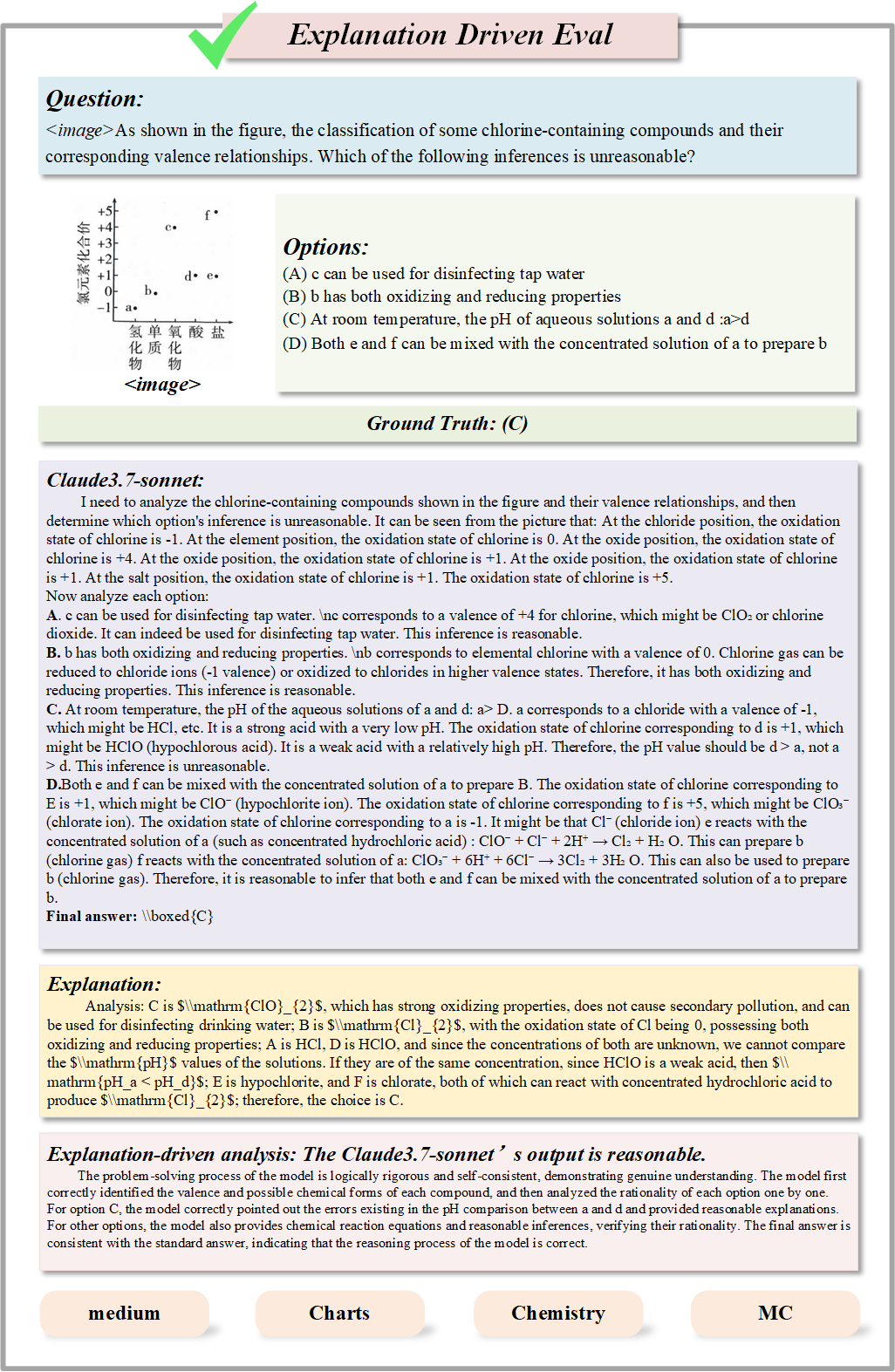}
		\caption{Successful reasoning example of Claude3.7-sonnet.}
		\label{exp_r1}
	\end{figure}

\begin{figure}[p]
		\centering
		\includegraphics[width=\textwidth]{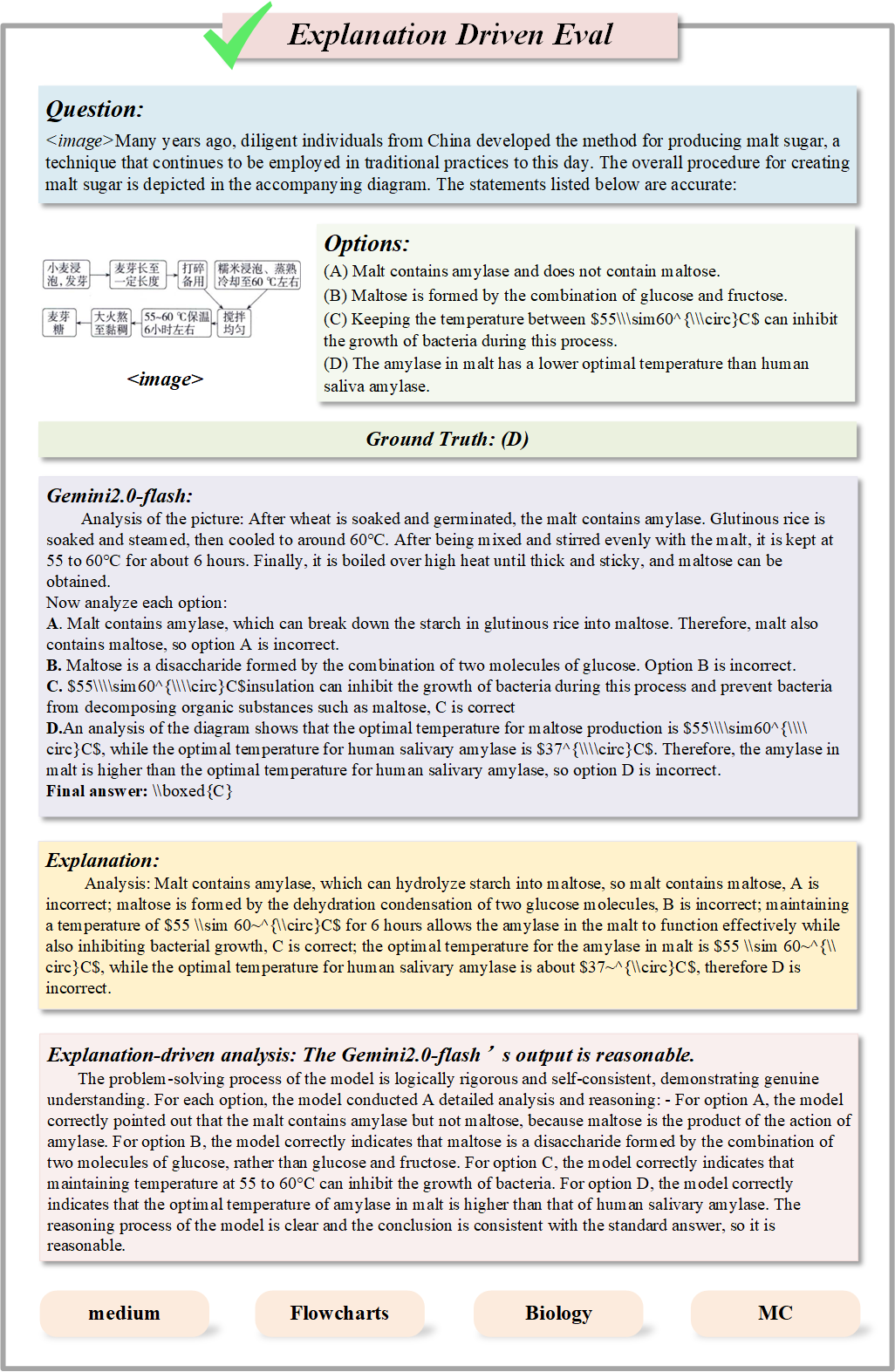}
		\caption{Successful reasoning example of Gemini2.0-flash.}
		\label{exp_r2}
	\end{figure}

\begin{figure}[p]
		\centering
		\includegraphics[width=\textwidth]{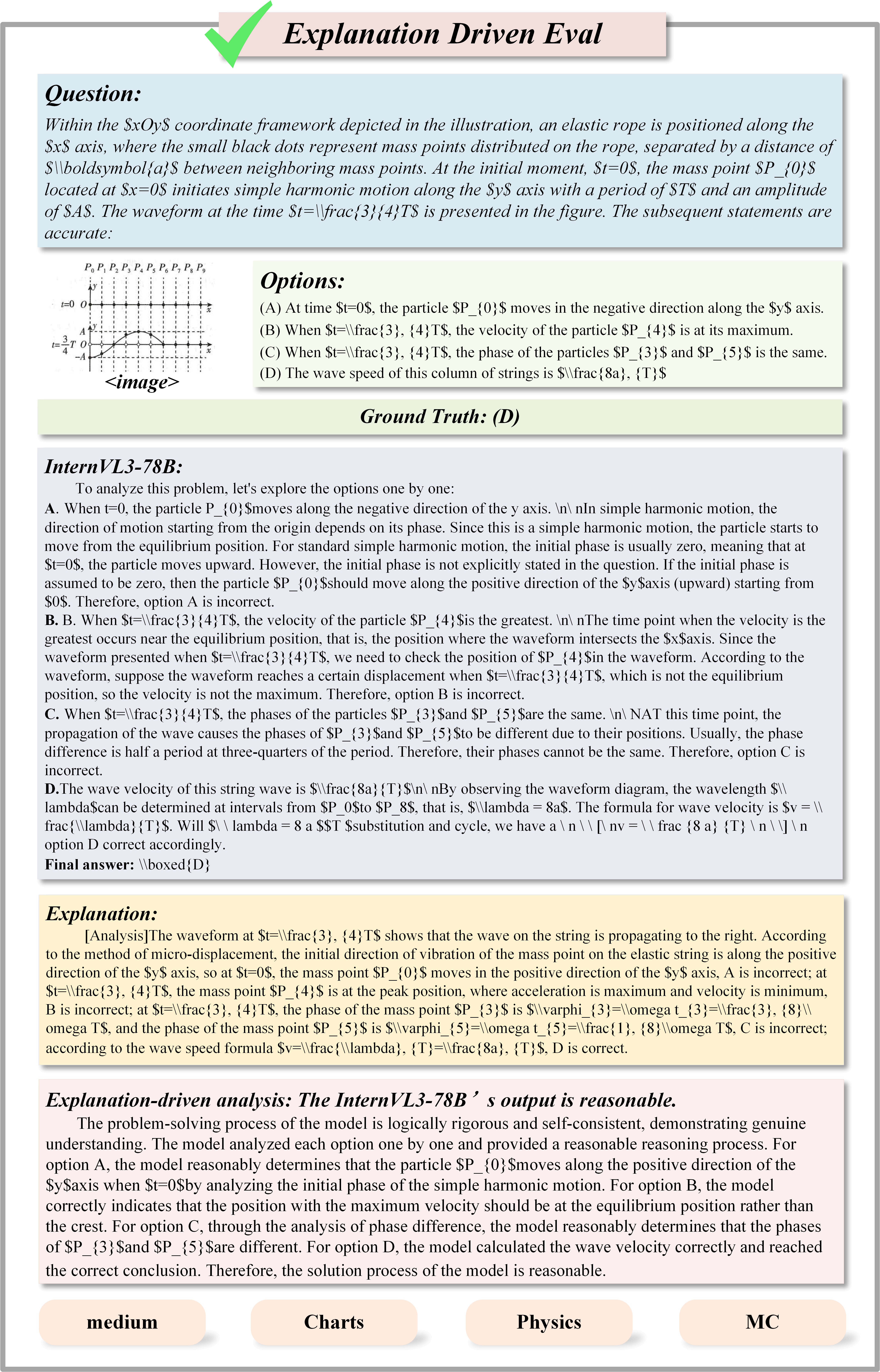}
		\caption{Successful reasoning example of InternVL3-78B.}
		\label{exp_r3}
	\end{figure}

\begin{figure}[p]
		\centering
		\includegraphics[width=\textwidth]{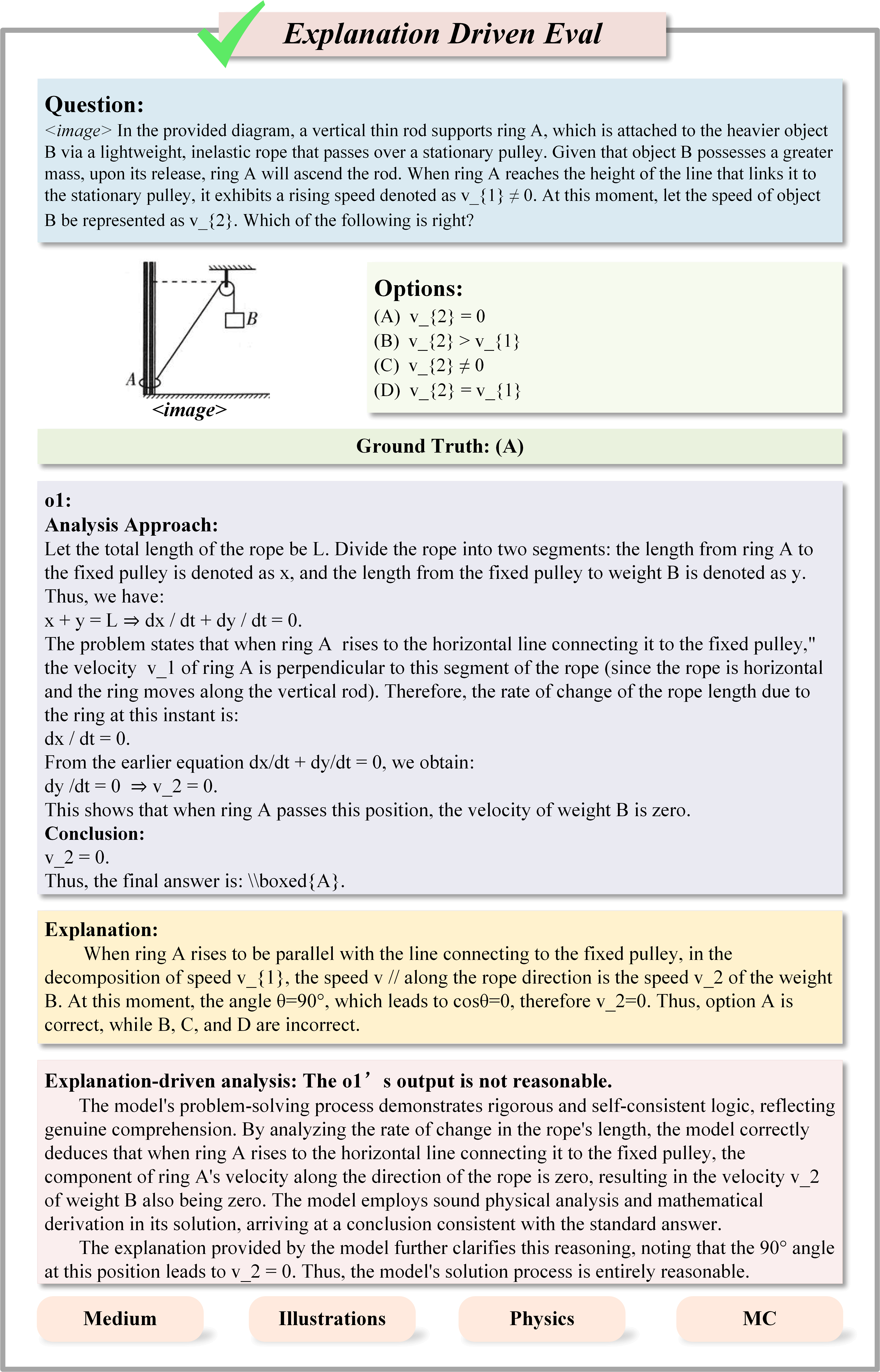}
		\caption{Successful reasoning example of o1.}
		\label{exp_r4}
	\end{figure}

\begin{figure}[p]
		\centering
		\includegraphics[width=\textwidth]{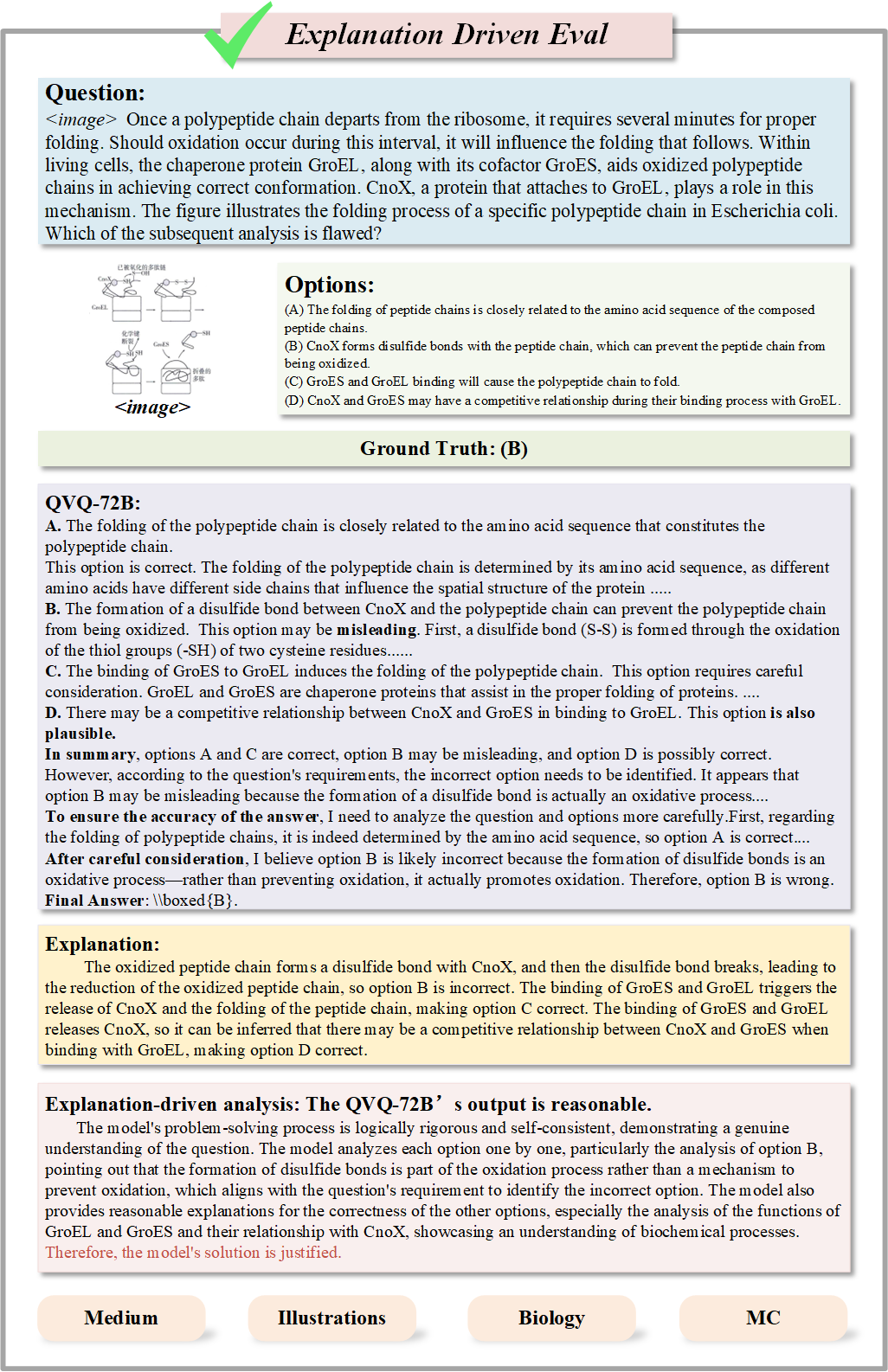}
		\caption{Successful reasoning example of QVQ-72B.}
		\label{exp_r5}
	\end{figure}

\begin{figure}[p]
		\centering
		\includegraphics[width=\textwidth]{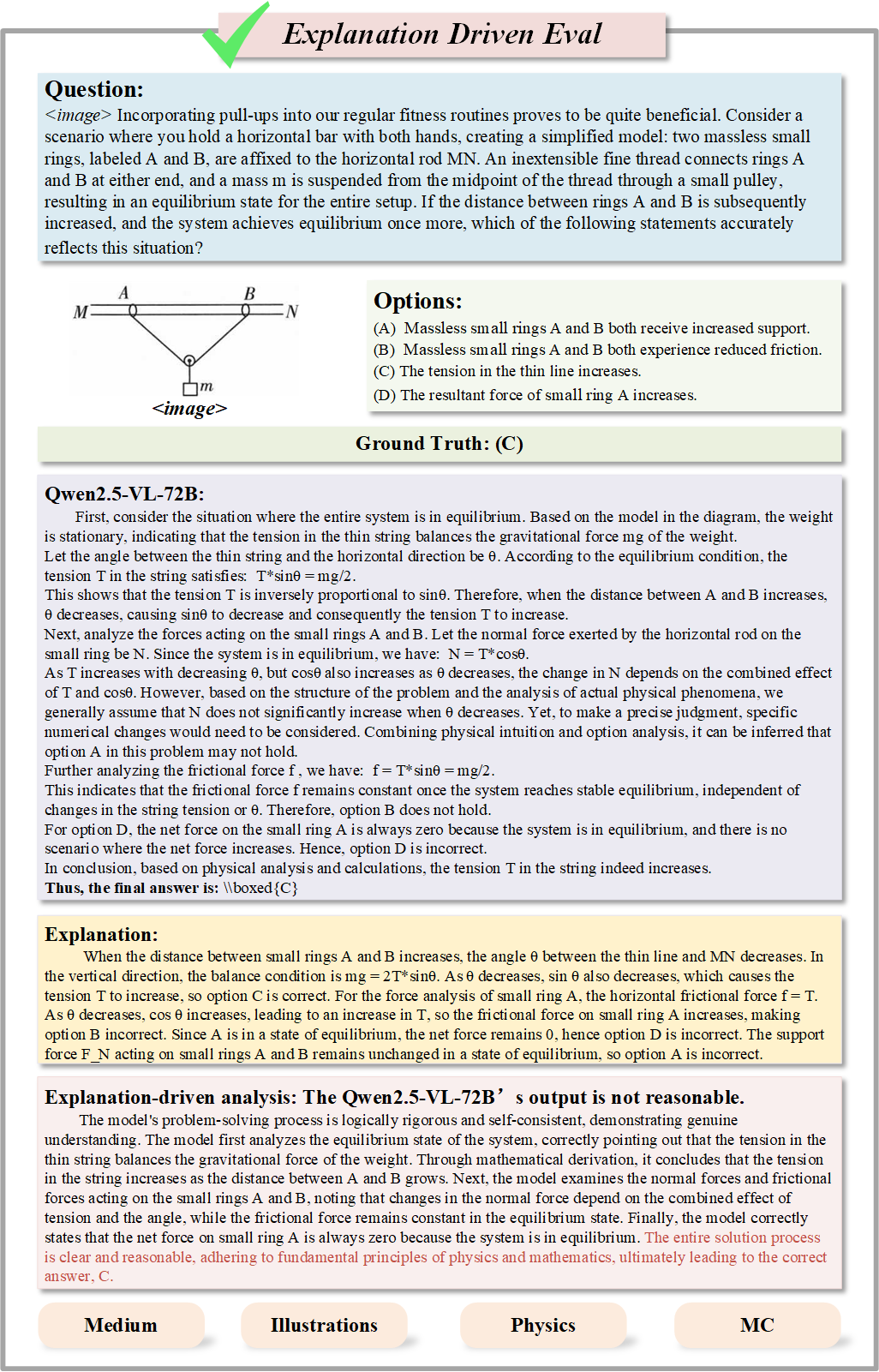}
		\caption{Successful reasoning example of Qwen2.5-72B.}
		\label{exp_r6}
	\end{figure}

\begin{figure}[p]
		\centering
		\includegraphics[width=\textwidth]{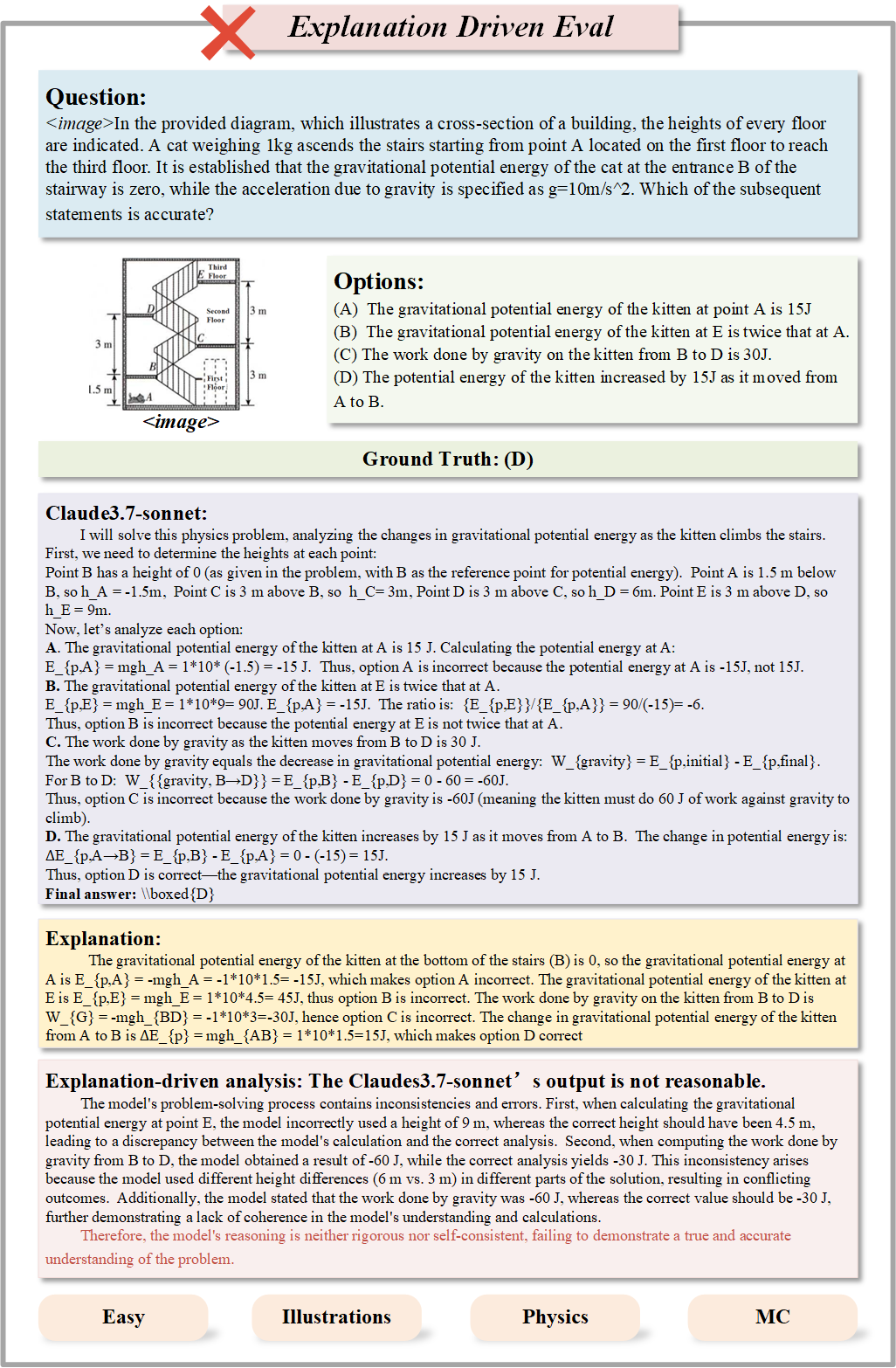}
		\caption{Incorrect reasoning example of Claude3.7-sonnet.}
		\label{exp_e1}
	\end{figure}

\begin{figure}[p]
		\centering
		\includegraphics[width=\textwidth]{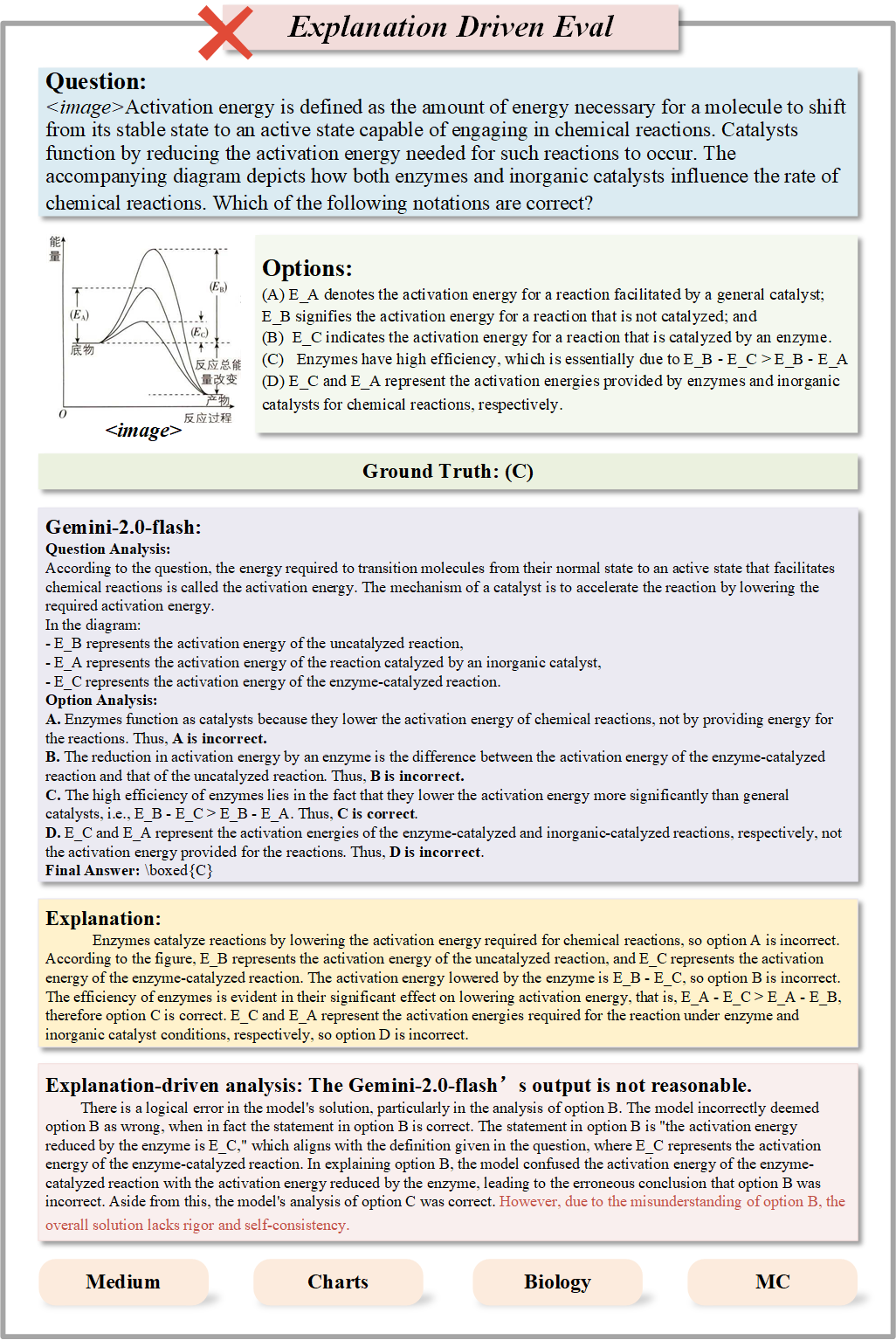}
		\caption{Incorrect reasoning example of Gemini2.0-flash.}
		\label{exp_e2}
	\end{figure}

\begin{figure}[p]
		\centering
		\includegraphics[width=\textwidth]{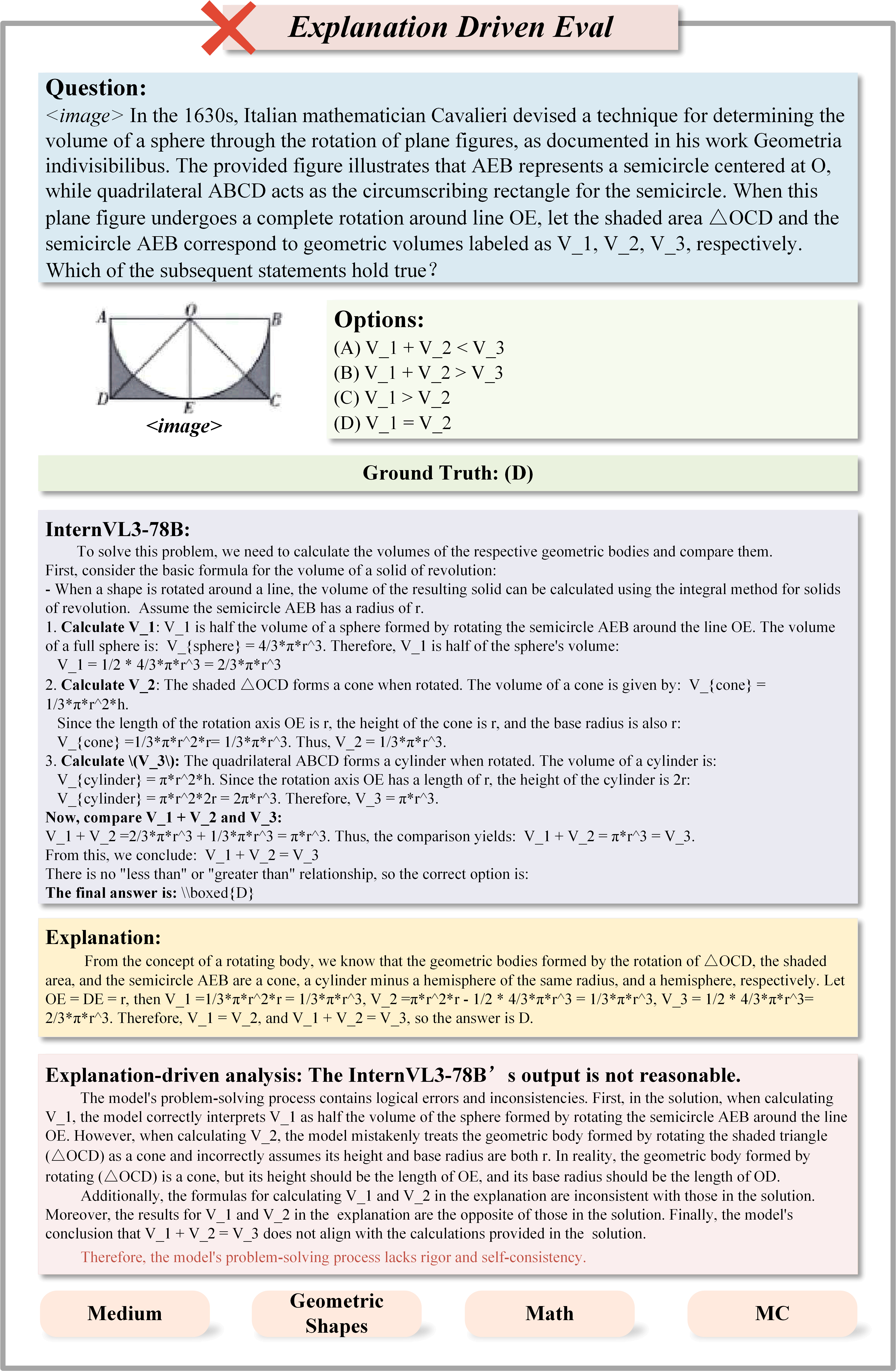}
		\caption{Incorrect reasoning example of InternVL3-78B.}
		\label{exp_e3}
	\end{figure}

\begin{figure}[p]
		\centering
		\includegraphics[width=\textwidth]{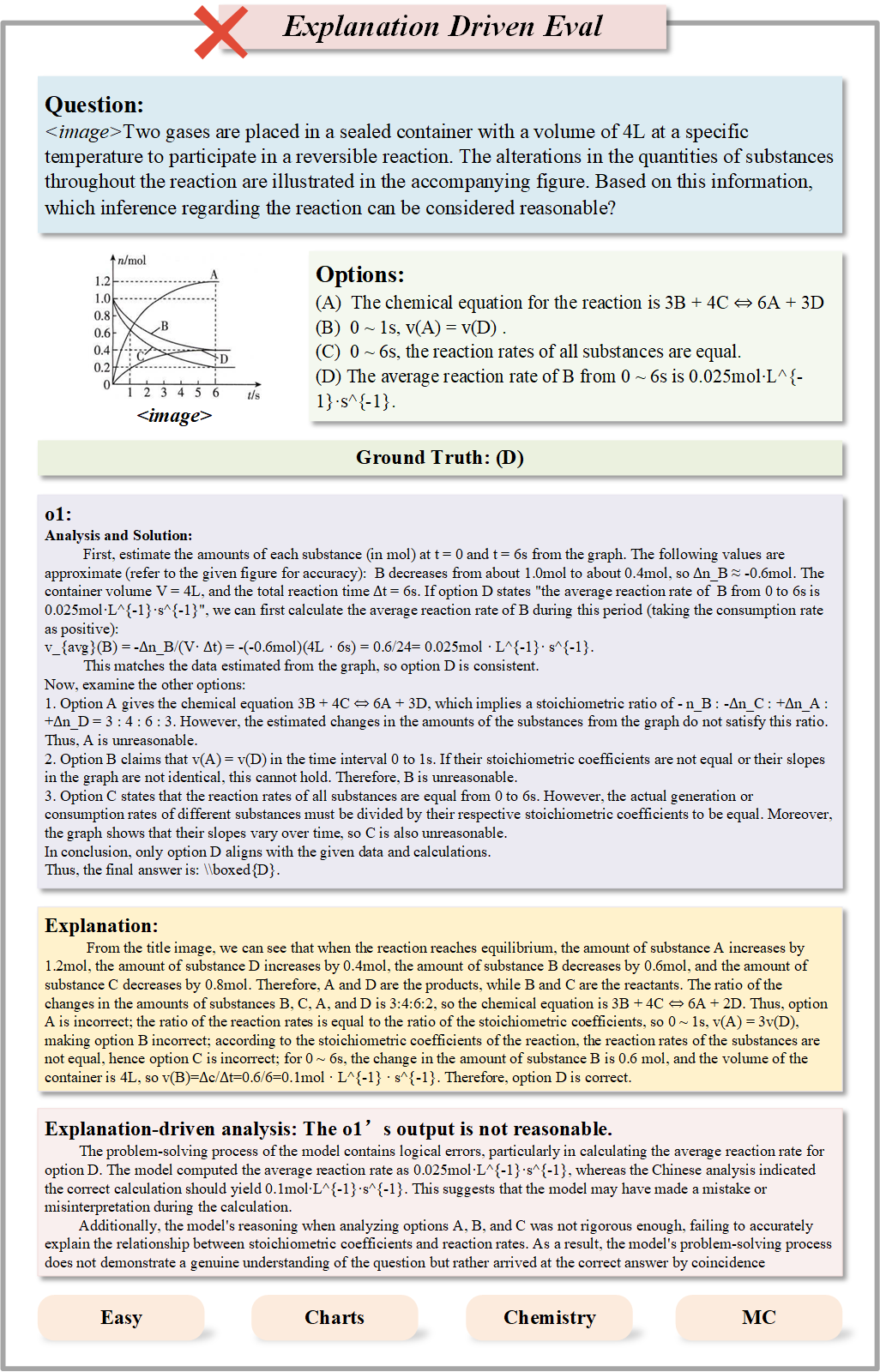}
		\caption{Incorrect reasoning example of o1.}
		\label{exp_e4}
	\end{figure}

\begin{figure}[p]
		\centering
		\includegraphics[width=\textwidth]{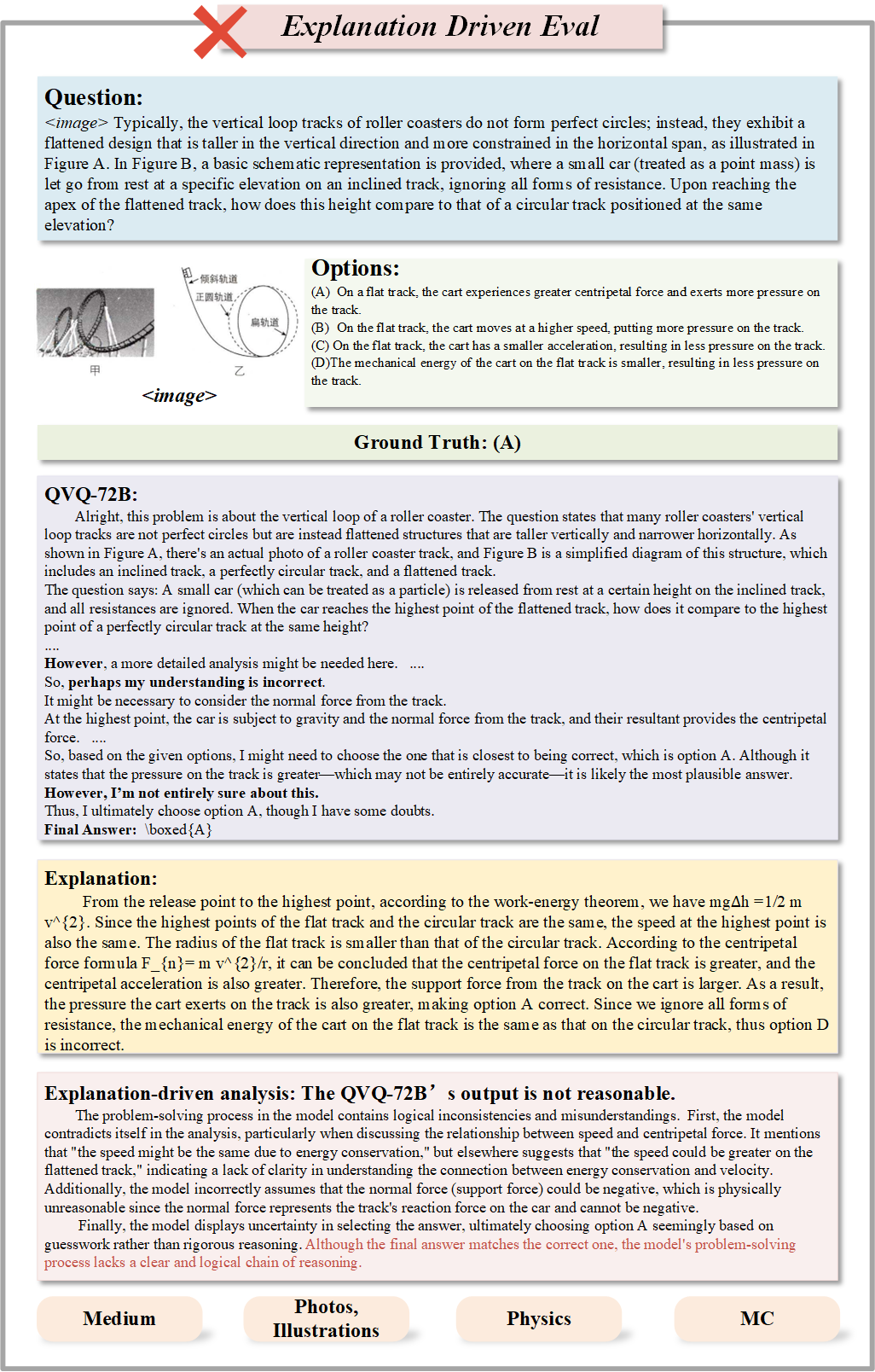}
		\caption{Incorrect reasoning of of QVQ-72B.}
		\label{exp_e5}
	\end{figure}

\begin{figure}[p]
		\centering
		\includegraphics[width=\textwidth]{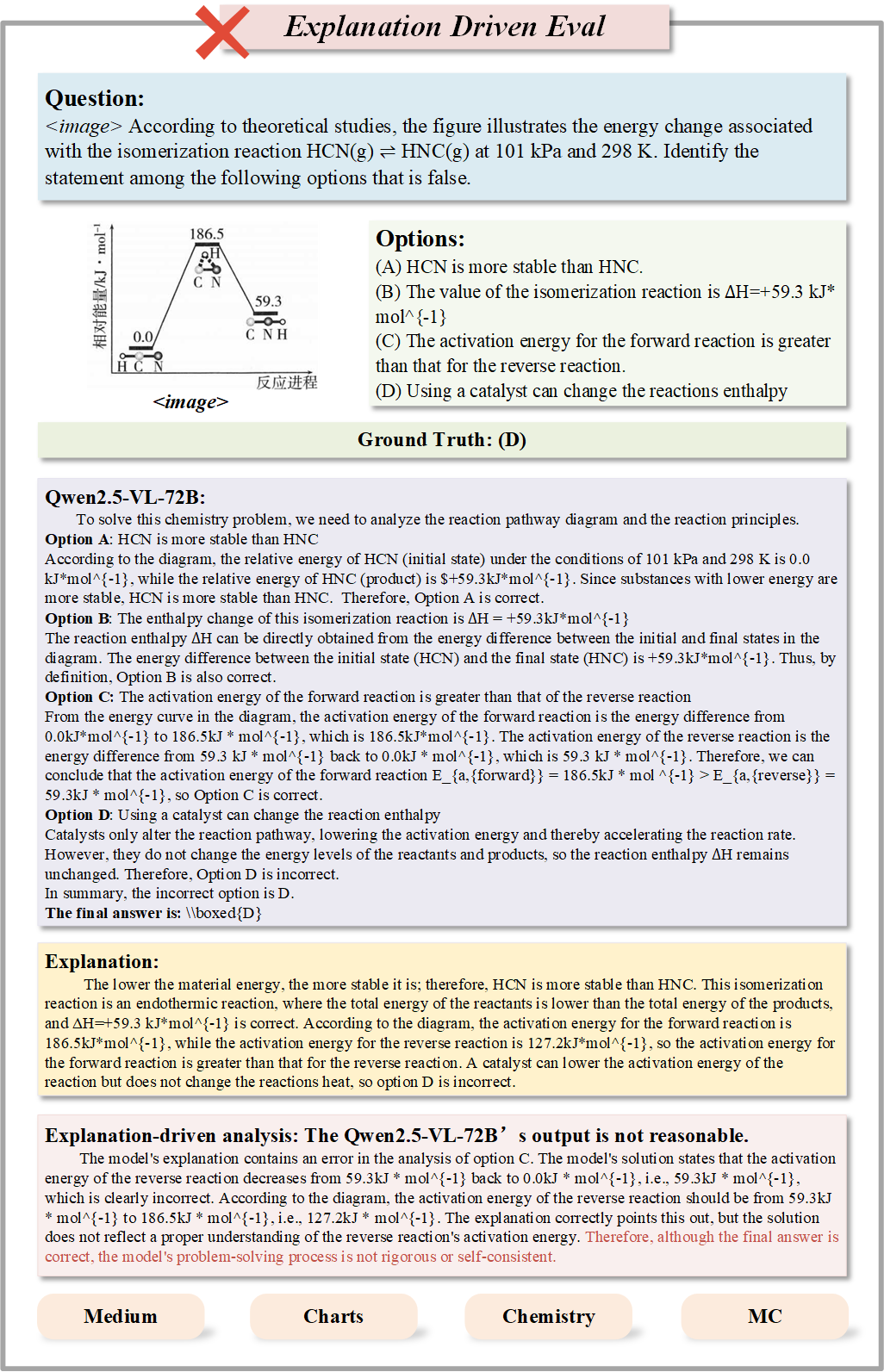}
		\caption{Incorrect reasoning example of Qwen2.5-72B.}
		\label{exp_e6}
	\end{figure}

	We focus on correct predictions and assess whether the explanation plausibly reflects the underlying image and question context. Cases where the reasoning is illogical, hallucinated, or merely pattern-based are labeled as lucky guesses. Conversely, coherent and grounded explanations are treated as valid reasoning. 
	Examples in Figures~\ref{exp_r1}–\ref{exp_r6} demonstrate cases with strong reasoning consistency, while Figures~\ref{exp_e1}–\ref{exp_e6} showcase flawed reasoning chains behind otherwise correct answers.
	
	These analyses reveal that even high-performing models sometimes rely on superficial correlations rather than deep understanding, reinforcing the need for explanation-aware benchmarks in multimodal evaluation.
	
\end{document}